\documentclass{applemlr}

\usepackage{amsmath}
\usepackage{enumerate}
\usepackage{algorithm}
\usepackage{algpseudocode}
\usepackage{amsfonts}
\usepackage{amsthm}
\usepackage{cleveref}
\usepackage{diagbox}
\usepackage{colortbl}
\usepackage{amssymb}
\usepackage{xspace}
\usepackage{wrapfig}
\usepackage{adjustbox}
\usepackage{tabularx}
\usepackage{booktabs}
\usepackage{mathtools}
\usepackage{tikz}
\usepackage{enumitem}
\usepackage{silence}
\usepackage{dsfont}
\usepackage[table]{xcolor}
\usepackage[dvipsnames]{xcolor}
\usepackage{multirow}
\usepackage{makecell}
\usepackage{xfakebold}

\usepackage{amsmath,amsfonts,bm}

\def\eqref#1{equation~\ref{#1}}

\def\1{\bm{1}}

\DeclareMathAlphabet{\mathsfit}{\encodingdefault}{\sfdefault}{m}{sl}
\SetMathAlphabet{\mathsfit}{bold}{\encodingdefault}{\sfdefault}{bx}{n}

\definecolor{textgray}{HTML}{6E6E73}
\usetikzlibrary{positioning, calc}
\usetikzlibrary{decorations.pathmorphing}

\makeatletter
\patchcmd{\wrong@fontshape}{\@gobbletwo}{}{}{}
\makeatother
\numberwithin{equation}{section}
\makeatletter
\AtBeginDocument{
  \urlstyle{sf}
  
}
\makeatother

\definecolor{light}{RGB}{125, 125, 125}
\crefname{tcb@cnt@pbox}{code}{code}
\Crefname{tcb@cnt@pbox}{Code}{Code}
\crefname{assumption}{assumption}{assumption}
\Crefname{assumption}{Assumption}{Assumptions}

\newtcolorbox[auto counter]{pbox}[2][]{
  colback=white,
  title=Code~\thetcbcounter: #2,
  #1,fonttitle=\sffamily,
  fontupper=\sffamily,
  arc=2pt,
  colframe=bgcolor,
  coltitle=fgcolor,
  colbacktitle=bgcolor,
  toptitle=0.25cm,
  bottomtitle=0.125cm
}

\makeatletter
\newcommand\applefootnote[1]{%
  \begingroup
  \renewcommand\thefootnote{}%
  \renewcommand\@makefntext[1]{\noindent##1}%
  \footnote{#1}%
  \addtocounter{footnote}{-1}%
  \endgroup
}
\makeatother

\definecolor{cverbbg}{gray}{0.90}

\usepackage{natbib}
\usepackage{microtype}
\usepackage{hyperref}
\usepackage{url}
\usepackage{booktabs}
\usepackage{graphicx}
\usepackage{subcaption}
\usepackage{xcolor}
\usepackage{array}
\usepackage{multirow}
\usepackage{float}
\usepackage{subcaption}
\usepackage{textcomp}
\usepackage{colortbl}
\usepackage{amsmath}
\usepackage{amssymb}
\usepackage{makecell}    
\usepackage{soul}
\usepackage{wrapfig}
\definecolor{rowhighlight}{RGB}{255, 249, 222}    %
\definecolor{colhighlight}{RGB}{232, 237, 248}    %
\definecolor{cellhighlight}{RGB}{235, 233, 210}   %

\newcommand{\hlrow}[1]{{\sethlcolor{rowhighlight}\hl{#1}}}

\usepackage{lineno}

\definecolor{darkblue}{rgb}{0, 0, 0.5}
\hypersetup{colorlinks=true, citecolor=darkblue, linkcolor=darkblue, urlcolor=darkblue}

\makeatletter
\newcounter{finding}
\newcommand{\finding}[1]{%
  \refstepcounter{finding}%
  \paragraph{Finding \thefinding: #1}%
  \def\@currentlabel{\thefinding}%
}
\makeatother

\title{What do your logits know?}

\author[*]{Masha Fedzechkina}
\author[*]{Eleonora Gualdoni}
\author{Rita Ramos}
\author[*]{Sinead Williamson}

\begingroup
\renewcommand{\thefootnote}{}
\footnotetext{This work is a team effort. * indicates a core author.}
\endgroup

\affiliation{Apple}

\abstract{
Recent work has shown that probing model internals can reveal a wealth of information not apparent from the model generations. This poses a risk of unintentional or malicious information leakage, where model users are able to learn information that the model owner assumed was inaccessible. Using vision-language models as a testbed, we present the first systematic comparison of information retained at different representational levels as it is compressed from the rich information encoded in the residual stream through two natural bottlenecks: low-dimensional projections of the residual stream obtained using tuned lens, and the final top-$k$ logits most likely to impact model's answer. We show that even easily accessible bottlenecks defined by the model's top logit values can leak task-irrelevant information present in an image-based query, in some cases revealing as much information as direct projections of the full residual stream.
}

\metadata[Correspondence]{\sffamily Masha Fedzechkina: \url{mfedzechkina@apple.com}; Sinead Williamson: \url{sa_williamson@apple.com}}
\date{\sffamily\today}

\begin{document}

\maketitle

\section{Introduction}

When making a decision or answering a question, a model should consider all relevant information at hand and ignore all irrelevant information. This is the key to the information bottleneck principle  \citep[IB,][]{tishby1999information,tishby2015deeplearninginformationbottleneck} %
as excluding relevant information will lead to suboptimal decisions, while basing a decision on irrelevant information (that may be spuriously correlated with the task at hand) opens the door to generalization failures and undesirable biases. Moreover, understanding what information is encoded in a model’s output allows us to better understand the risk of information leakage: if a model’s output contains information about a user’s sensitive data, then that data may be vulnerable to malicious attacks. 

\begin{figure}[h]
   \centering
    \includegraphics[height=4cm]{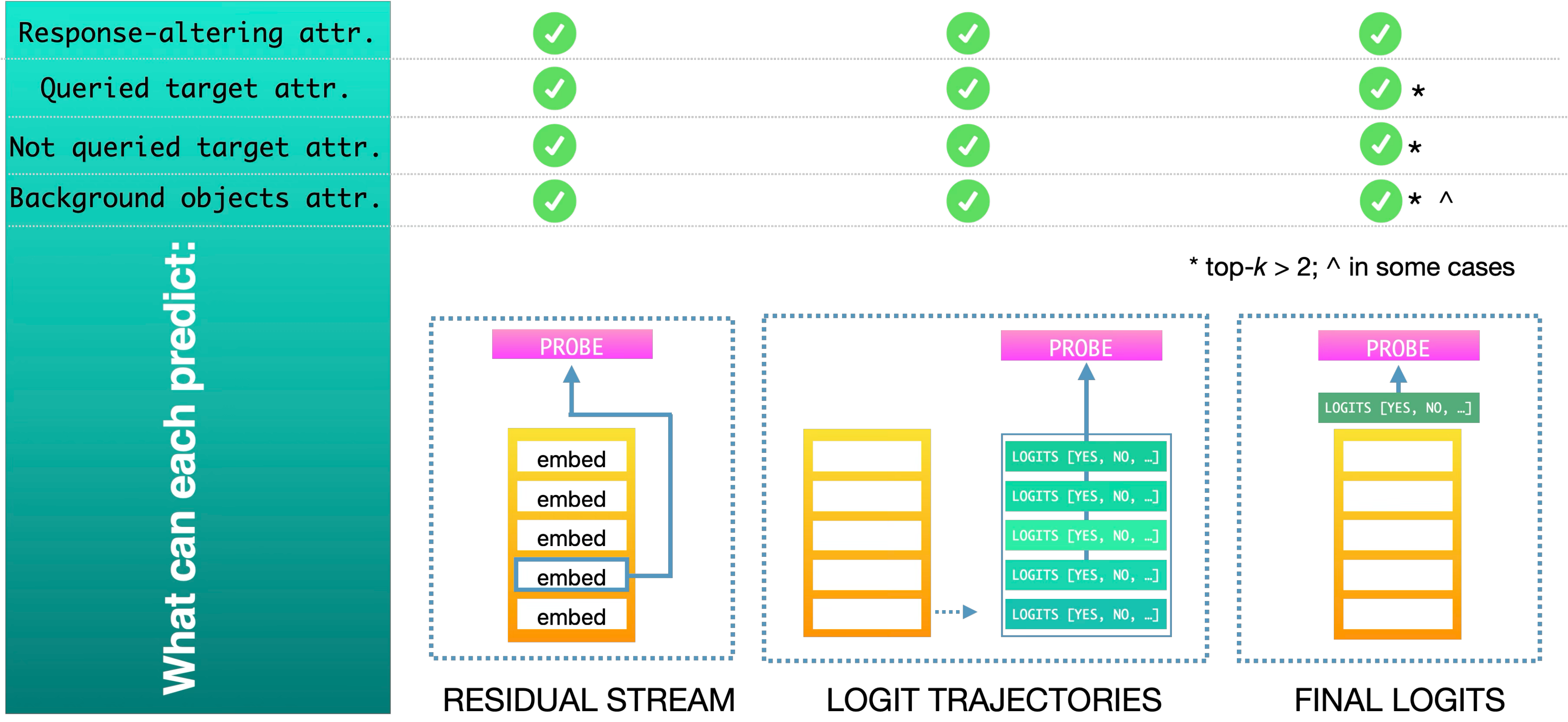}
    \caption{High-level summary of our findings. Response-altering, target information (queried and non-queried), and in many cases the information about the background objects can be reliably decoded in the final logits.}
    \label{fig:teaser}
\end{figure}

Prior work has provided suggestive evidence, which our findings  further support, that the residual stream captures a large amount of information contained in a query and extracted from the LLM’s training data. For instance, work on probing classifiers showing that the residual stream encodes rich representations including part-of-speech information, facts, or world knowledge, with different types of information peaking in different layers \citep{tenney2019bertrediscoversclassicalnlp, meng2023locatingeditingfactualassociations}. Probing work has also demonstrated that the residual stream can contain information that \textit{contradicts} the model's final answer \citep[e.g., the hidden states at some layers may contain the true answer even when the model hallucinates;][]{halawi2024overthinkingtruthunderstandinglanguage}. %

As we move towards lower-dimensional representations (such as restrictions of the output distribution to just the top-$k$ logits), some of this information must necessarily be discarded -- a response to a yes/no question cannot possibly contain the rich information encoded in the residual stream. This raises the following questions that are the focus of the current work: At what point in processing is this extraneous information discarded? And does any of it ``survive'' to be decodable in the final logits?

Inspired by the IB, we present, to our knowledge, the first systematic comparison of information retained at different representational levels of vision-language models (VLMs) to understand how much, and what, information is retained in the output as the comprehensive information stored in the residual stream is compressed during hierarchical processing. Specifically, we consider the hidden states that make up the residual stream that are expected to contain rich representations of the scene based on prior work \citep{conneau-etal-2018-cram, belinkov2019analysismethodsneurallanguage}, and two bottlenecks: low-dimensional projections of the residual stream obtained using tuned lens \citep{belrose2023eliciting}, which have been claimed to retain an intermediate amount of information between the hidden states and the final generations \citep{cywinski2025eliciting}, and the final top-$k$ logits, most likely to impact the decisions under stochastic generation. %
 
In our setup, we present a VLM with an image and ask it ``is there a \texttt{\textlangle category\textrangle} in the image? Reply in one word''. This offers a scenario where the query contains a rich source of information, much of which is irrelevant to the task, and allows us to explore what type of information remains recoverable as it is progressively compressed between the high-dimensional residual stream and the single-token response.

Our findings are as follows (see Figure~\ref{fig:teaser} for a high-level summary):  
\begin{enumerate}
    \item The residual stream can be seen as an ``oracle'': it is highly predictive of all aspects of a query image, regardless of whether they are queried or not. Conversely, the top two final logits (i.e., the logits associated with the tokens ``Yes" and ``No") contain primarily response-altering information (such as, whether the image has been corrupted by noise).
    \item As we increase the number of final logits observed, we see more leakage of non-queried information. For instance, we are able to reliably predict information about the target that was never mentioned in the query (e.g., the material and size of the sphere if we ask ``is there a blue sphere?'') and, to some degree, information about other objects in the image.
    \item Logit trajectories have previously been shown to leak unintended information from the residual stream to the user, providing a security vulnerability \citep{cywinski2025eliciting,belrose2023eliciting}. We find that, in general, the top-$k$ logits leak less non-target information (e.g., properties of background objects) than size-matched trajectories. However, they leak comparable amounts of target-related information. 
    This suggests that final logits, which are often directly accessible by the user or inferrable via repeated sampling, are almost as vulnerable to information-gathering attacks as representations of the full model internals. The observed leakage has implications for several areas of machine learning, which we discuss in Section~\ref{sec:discussion}.
\end{enumerate}

\section{Related Work}

\subsection{Information bottleneck}

The IB principle \citep{tishby1999information} defines an optimal representation as one that retains only information relevant to the target, while discarding irrelevant variation. The IB provides a general framework for representation learning across different kinds of systems. For instance, this perspective has been applied to human communication, where semantic structure appears to reflect such efficiency trade-off \citep{Zaslavsky2018}. In deep learning, \cite{tishby2015deeplearninginformationbottleneck} and \cite{shwartzziv2017openingblackboxdeep} interpret neural network training through this lens, proposing that deep learning models undergo a compression phase during training in which layers progressively discard input information that is not predictive of the output.

In practice, a model's representations may differ from such an optimal bottleneck in two ways: they may discard information that \textit{is} relevant to the prediction, and they may retain information that \textit{is not} relevant to the prediction. In the context of language modeling, \cite{yang2018breakingsoftmaxbottleneckhighrank} identified a \textit{softmax bottleneck}: the final log-linear layer imposes a low-rank constraint on the model's output distribution, limiting its expressiveness. However, \citet{basrisoftmax} showed that this restriction may not matter in practice: for reasonable values of $k$, the rank constraint does not limit the expressiveness of the top-$k$ logits. %

Equally concerningly, the output of an LLM may contain information that is not directly relevant to the prediction. This may be due to spurious correlations in the training data \citep{geirhos2020shortcut,zhou2024explore} or biases introduced via post-training \citep{santurkar2023whose}. However, it may also be due to a failure of compression: convergence to an optimal bottleneck is not guaranteed \citep{saxe2019information}, and recent work indicates that even late-layer representations contain very rich information \citep{gurnee2023language,burns2024discoveringlatentknowledgelanguage}. In particular, the architecture of transformers makes use of residual connections, which tend to inhibit the compression of information as it is processed through the network \citep{orhan2017skip,behrmann2019invertible,elhage2021mathematical}. Moreover, IB defines optimality with respect to a specific target but general-purpose LLMs need to support arbitrary targets (e.g., QA, follow-up questions, captioning, etc.) that are not necessarily known in advance. Together, these considerations leave open the question of what the target of the IB is for general-purpose LLMs.

\subsection{Understanding the information in the hidden states: Probing classifiers}
Probing classifiers have been the primary tool for understanding the type of information encoded in the model's internal representations. This approach involves training a light-weight (typically linear) classifier that maps hidden-state representations (typically, the residual stream, but sometimes other representations such as attention weights) to linguistic or visual properties of interest, with the classifier accuracy taken as a proxy for the accessibility of the information in the model's representation \citep{conneau-etal-2018-cram, belinkov2019analysismethodsneurallanguage}. Crucially, probing measures information availability in a representation, not causal use, a distinction well-established in the literature \citep{belinkov_probing_survey}. This work has revealed that the model's residual stream contains information that goes beyond what is contained in the next-token predictions. For instance, probing classifiers have been shown to recover syntactic parse trees \citep{hewitt-manning-2019-structural}, factual associations and world knowledge \citep{burns2024discoveringlatentknowledgelanguage, petroni-etal-2019-language}, truthfulness \citep{marks2024geometrytruthemergentlinear} and correct targets for hallucinated output \citep{orgad2025llmsknowshowintrinsic}. In sum, this line of work has established that the model's residual stream is informationally rich and redundant compared to the model's output. This raises the question of how much and what type of the information contained in the hidden states survives the projection into the logit space, from which the decisions are ultimately made, which is the focus of the current work.

\subsection{Peering into the residual stream: Logit lens and derivatives} A complementary approach has sought to understand the dynamics of information processing in transformers by directly examining how the model's predictions evolve across layers. Logit lens, introduced by \citet{nostalgebraist2020}, is a simple but powerful technique that projects the model's residual stream into the logit space via the unembedding matrix, providing pseudo-logits (i.e., the model's prediction for the output at that layer, assuming nothing more is added to the residual stream). When applied to each intermediate layer, this technique provides a trajectory of pseudo-logits that reflects how the model's predictions evolve from diffuse early predictions to sharp final outputs. \citet{belrose2023eliciting} refined the approach with tuned lens, which learns an affine transformation into the final layer before the projection into the unembedding matrix, and provides better-calibrated predictions for the intermediate layers. Projections obtained using these methods have been shown to contain information about facts that the LLM has been explicitly instructed \textit{not} to reveal \citep{cywinski2025eliciting,belrose2023eliciting}, suggesting that they retain at least some of the information from the residual stream that does not appear in the final logits. These findings motivate us to consider logit trajectories as a natural representational layer between the residual stream and the output logits, since they demonstrate that these trajectories are \textit{not} a minimal information bottleneck. We compare the information found in tuned lens projections with both the full residual stream, and other, more easily accessible, candidate representational levels of comparable dimensionality in the form of final-layer logits.

\section{Method}
Our work is conceptually motivated by the idea of the IB and the observation (from prior work) that a model's residual stream contains a significant amount of information that is seemingly irrelevant to the final next-token prediction. %
Our goal is to explore what information is retained, and what is discarded, at different representational levels. We design a series of experiments based on visual question answering. This allows us to specify an input with a rich amount of information, much of which is not queried. We verify, using probes, that both queried and not-queried information can be extracted from the full residual stream. We then consider two forms of lower-dimensional representation: 1) the trajectory of the expected logits obtained via tuned lens and 2) final-layer logits. These can be easily accessed in either a white-box (i.e., full access to model internals) or a grey-box (i.e., access to output logits) manner. We probe what information is decodable in the representation at these informational levels.

\subsection{The information bottleneck task}
\label{sec:task}
We are inspired by the IB as a controlled experimental scenario: it creates a maximally narrow channel that allows us to quantify how much and what information about the input to a model is decodable at various representation levels of the model's output. The input in each case is a simple binary question: We present an image, and ask the model ``Is there a \texttt{\textlangle category\textrangle} in the image? Reply in one word". The output under consideration is the first generated token (since we have explicitly requested the model to reply in one word). 

For each query, we consider several forms of information about the image:
\begin{itemize}
    \item \textbf{Response-altering attributes} such as noise applied to the image: These are explicit manipulations we apply to an image to alter the model's response (see Appendix~\ref{app:clevr_acc_noise_level},  ~\ref{app:clevr_acc_noise_level_images} and ~\ref{app:mscoco_noise_sweep} for details of the empirical investigation).
    \item \textbf{Target attributes}: attributes of the target \texttt{\textlangle category\textrangle}, such as color or material that can be queried or not. %
    \item \textbf{Background object attributes}: attributes of objects in the image that are \textit{not} the target \texttt{\textlangle category\textrangle}, and are never explicitly queried.
\end{itemize}

We train probes to predict these attributes from various representational levels of the next-token output (see Section~\ref{sec:probes} and Appendix~\ref{app:probes} for details).

\subsection{Dataset construction}

  \begin{figure}[h]
      \centering
      \setlength{\tabcolsep}{1pt}
      \renewcommand{\arraystretch}{0.8}
      \begin{tabular}{@{}c ccccc@{}}
          \rotatebox{90}{\small Gaussian noise} &
          \includegraphics[width=0.19\textwidth]{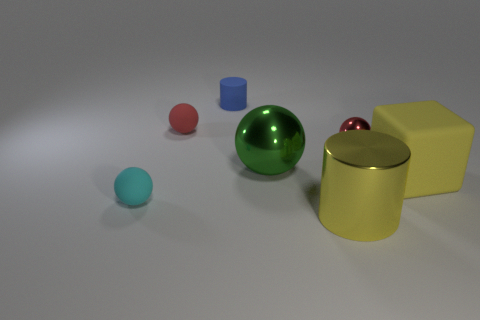} &
          \includegraphics[width=0.19\textwidth]{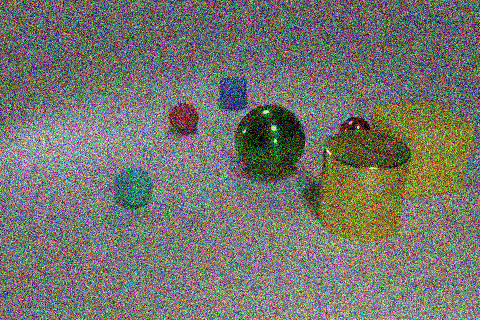} &
          \includegraphics[width=0.19\textwidth]{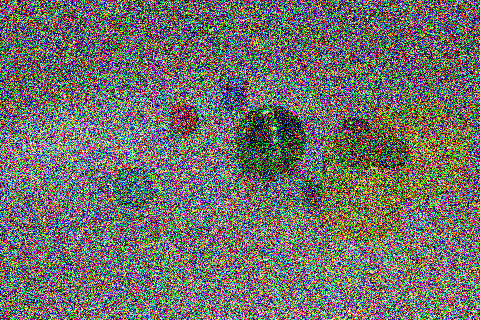} &
          \includegraphics[width=0.19\textwidth]{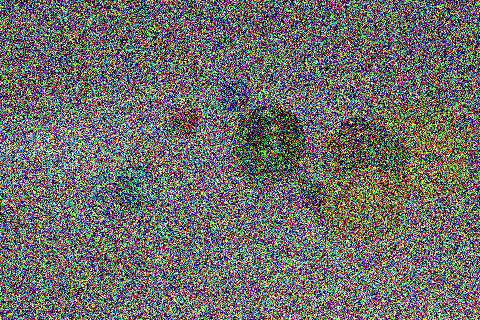} &
          \includegraphics[width=0.19\textwidth]{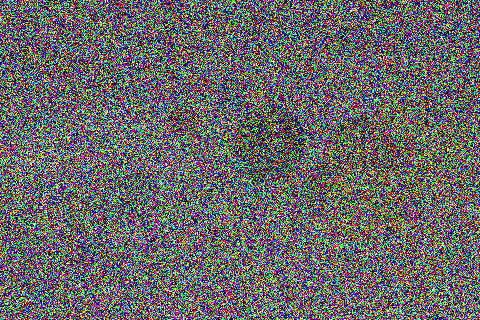} \\
          \rotatebox{90}{\small Motion blur} &
          \includegraphics[width=0.19\textwidth]{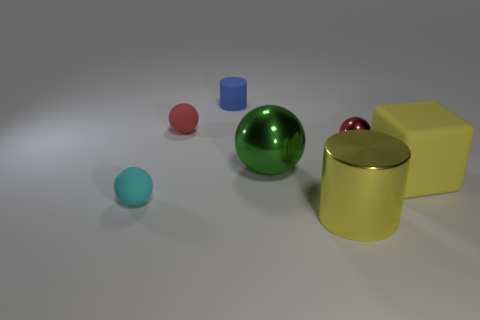} &
          \includegraphics[width=0.19\textwidth]{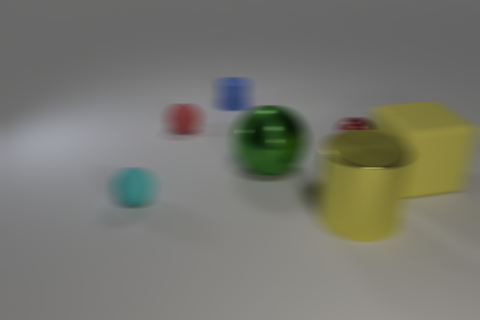} &
          \includegraphics[width=0.19\textwidth]{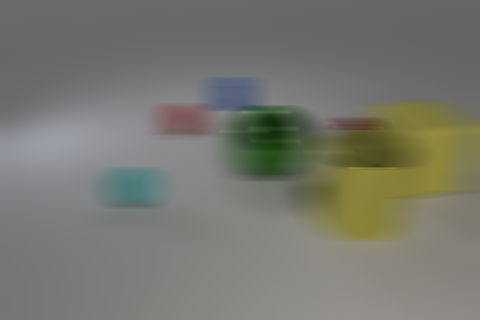} &
          \includegraphics[width=0.19\textwidth]{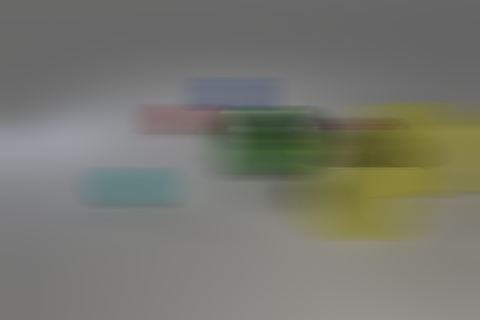} &
          \includegraphics[width=0.19\textwidth]{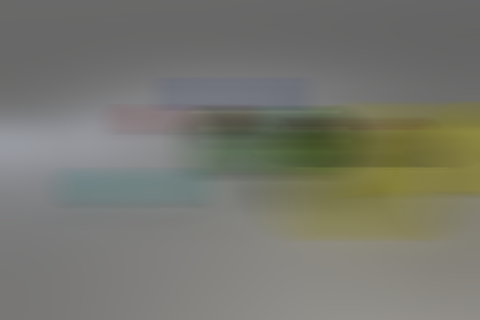} \\
          \rotatebox{90}{\small Glass blur} &
          \includegraphics[width=0.19\textwidth]{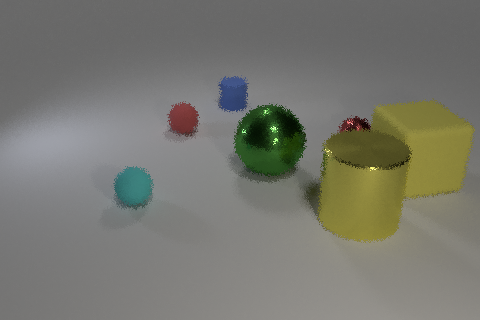} &
          \includegraphics[width=0.19\textwidth]{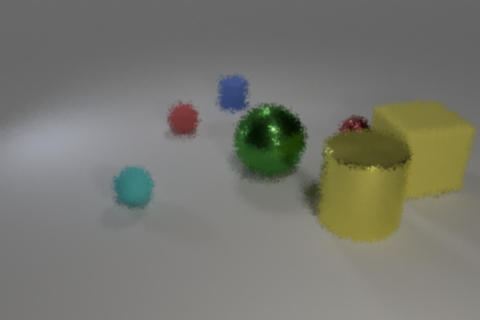} &
          \includegraphics[width=0.19\textwidth]{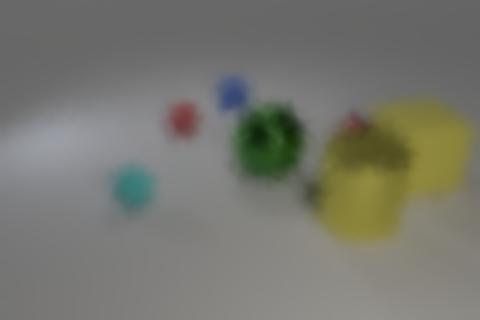} &
          \includegraphics[width=0.19\textwidth]{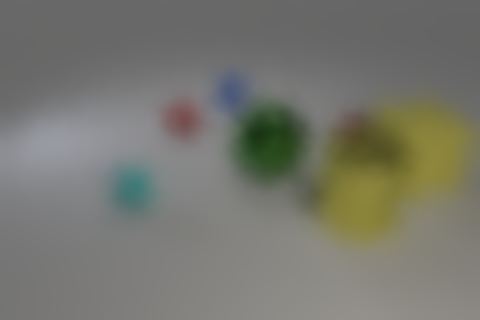} &
          \includegraphics[width=0.19\textwidth]{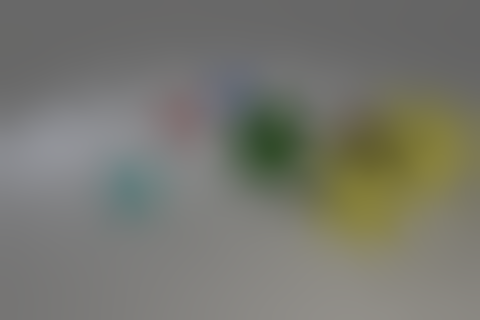} \\
      \end{tabular}
      \caption{Examples of three noise types applied at increasing strength to sample images in the CLEVR dataset (left to right). %
      Top
  row: Gaussian noise in steps. Middle row: motion blur. Bottom row: glass blur. For MSCOCO examples, see Appendix~\ref{app:mscoco_noise}}.
      \label{fig:noise_examples}
  \end{figure}

Our experiments are conducted on two datasets: a synthetic dataset with well-controlled images \cite[CLEVR;][]{clevr} and a widely-used dataset of naturalistic images \cite[MSCOCO;][]{mscoco}. CLEVR contains controlled scenes of three to ten objects with various attributes (shape: cube, sphere, and cylinder; size: small or large; materials: metal or rubber; and eight colors), making it easy to identify target-related and background-related attributes. We randomly sample 300 images for each number of objects in the scene (3-10) from the val set for a total of 2,400 images. For each image, we create various corrupted versions, by adding various noise types (Gaussian, glass, and motion noise) applied at increasing strength (see Appendix~\ref{app:mscoco_noise} and Figure \ref{fig:noise_examples}). For each image, we generate both ``positive'' queries (where the ground truth answer is ``yes'') and ``negative'' queries (where ground truth is ``no''). We generate a separate positive query for each object that appears in a given image, letting \texttt{\textlangle category\textrangle} be a minimal unique description for each target object, that refers to the object by its shape plus the minimal number of additional descriptors (e.g., color or material) required to uniquely identify the object within the image (e.g., \textit{gray rubber cube})\footnote{See Appendix \ref{app:clevr_description_length} for the statistics on the descriptions.}. We generate an equal number of ``negative'' prompts containing CLEVR descriptors of objects that do not appear in the image. 

For the MSCOCO val split, we generate a positive prompt for each of the 80 MSCOCO categories (e.g., ``cat" or ``bicycle") that appear in the image and a negative prompt where \texttt{\textlangle category\textrangle} is the category that most frequently co-occurs with the positive example but does not appear in the image. 

We pass the resulting queries through three high-performing VLMs: Qwen3-VL-8B-Instruct \citep[hereafter, Qwen3-VL-8B;][]{bai2023qwenvlversatilevisionlanguagemodel}, Llama-3.2-11B-Vision-Instruct \citep[hereafter, Llama-3.2-11B;][]{llama_clean}, and llava-v1.6-mistral-7b-hf \citep[hereafter, LLaVA-7B;][]{liu2023llava}.

\subsection{Representation levels}\label{sec:representation_levels}
For each model response, we collect the (greedily) generated text, plus several representational levels of the information passed to the first generated token:  (i) the representation of the residual stream at each layer of the model (hereafter referred to as hidden states); %
(ii) the $k$ highest-valued (final-layer\footnote{See Appendix~\ref{app:rep_levels} for details on final layer vs\. best layer performance for logits and hidden states.}) logits for the first generated token\footnote{We define the $k$ ``highest value'' logits as the logits with the $k$ largest maximal value over the training set; this corresponds to all tokens that \textit{ever} have probability above some threshold.} (in practice, $k$=2 corresponds to ``yes" and ``no"); (iii) the trajectory of expected logits for the first generated token, obtained by passing the full residual stream through tuned lens\footnote{See Appendix~\ref{app:tuned_lens_training} for the details on lens training and evaluations.}.  %

The hidden states can be seen as containing the maximum amount of information about the input. The generated text contains the minimal  
  amount of information (i.e., a single bit). The top-2 logits (denoted ``logits-2'' in our experiments) are the smallest representational level we consider: they can capture slightly more information than the generated tokens. As we increase the number of final-layer logits, we allow increasing amounts of information to be retained, enabling us to extrapolate between the amount of information in the final hidden state and that in the generated text. We sweep over multiples of the model depth $L$ (denoted ``logits-1L'', ``logits-2L'', \dots, ``logits-15L''; e.g., logits-2L refers to the top $2L$ logits where $L$ is model depth). We also consider all logits associated with variants of ``yes" and ``no" (e.g., ``Yes", ``NO"), denoted ``logits-all yes/no''. Additionally, we simulate a real-world scenario where we only have access to the top 20 logits (the amount exposed by many frontier model APIs) by creating a masked version of our dataset, where only the top 20 responses to a query are visible. We impute all other logit values to that of the smallest visible logit (i.e., the upper bound on their value), and use the resulting imputed dataset to train and evaluate probes (denoted ''imputed''; e.g., ''imputed-2L'').

The tuned lens trajectories provide an alternative extrapolation between hidden states and generated tokens. These trajectories are an affine transformation of the full residual stream, trained to predict the distribution over next token logits. As such, we would expect them to prioritize information that is relevant to the answer to our question. We consider the trajectories corresponding to the top-2 tokens (denoted ``trajectory-2'') and to all yes/no tokens (denoted ``trajectory-all yes/no''). 

\subsection{Probing representations for image information}
\label{sec:probes}
We construct probes to predict the following attributes:
\begin{itemize}
    \item \textbf{Response-altering attributes}: Noise level (for each of three noise types, Gaussian, glass and motion) and type of noise, trained and evaluated on both positive and negative queries using MSE. For CLEVR, we apply noise to the entire image; for MSCOCO we also apply the same noise to the target category or the context (everything but the target category) only since in natural images contextual support plays a role in scene understanding (e.g., images that contain a table are also likely to contain a chair \citep{li2023evaluatingobjecthallucinationlarge}), see Appendix~\ref{app:mscoco_context} for the illustration of the experiment setup and evidence of contextual support in MSCOCO.
    \item \textbf{Target attributes}: For CLEVR, we consider color, size, material, shape, and position within image (represented as a 3x3 grid), trained on only positive examples using cross-entropy and evaluated using accuracy. For MSCOCO, we include the position of the target on a 3x3 grid (if the image contains multiple targets, we use the location of the largest instance) and target object saliency (evaluated using MSE), which we obtain by providing GPT-5 with the MSCOCO labels for the image and asking it to judge how salient the object is on the scale from 0 to 100 based on the labels alone.
    \item \textbf{Background attributes}: For CLEVR, we include multi-hot representations of color, size, material, shape (where a 1 indicates at least one background object possesses that attribute) and the number of background objects in the image. All probes trained on only positive examples using cross-entropy; the number of background objects is evaluated using MAE and other attributes are evaluated using accuracy. For MSCOCO, we predict the presence of the category ''person'' in an image when this category is not queried \footnote{We selected person, as approximately 50\% of images include a person; other categories are seen substantially less frequently.} trained on a subset of the dataset that excludes the positive and negative queries involving ''person'' and evaluated using accuracy. 
\end{itemize}

Note that it only makes sense to consider positive queries when considering attributes of the target (since the target is not present in negative queries). %
We use linear probes to predict each attribute from the representational levels discussed in Section~\ref{sec:representation_levels}. We additionally fit lightweight CNN probes for tuned lens trajectories to account for the structured nature of these representations (details in Appendix~\ref{app:probes}). An 80/10/10 train/val/test split was generated based on image id, so that all queries based on a given image (and its noisy manipulations) belong to the same split.

\section{Results}\label{sec:results}
We use the three representational levels discussed in Section~\ref{sec:representation_levels} to predict scene attributes, leading to the following findings (for CLEVR, see Tables~\ref{tab:clevr_table1} and~\ref{tab:clevr_table2}; for MSCOCO, see Tables~\ref{tab:mscoco_table}). Overall, we observe that the patterns in the findings discussed below hold for both the synthetic CLEVR and naturalistic MSCOCO datasets.

\begin{table}[ht]
\centering
\resizebox{\textwidth}{!}{%
\begin{tabular}{llcccccc}
\hline
\multicolumn{2}{l}{} & \multicolumn{6}{c}{\textbf{Predicted Target}} \\
\cline{3-8}
\textbf{Model} & \textbf{Representation Type} & \cellcolor{colhighlight}\shortstack[t]{\textbf{noise level (MSE↓)}} & \cellcolor{colhighlight}\shortstack[t]{\textbf{noise type (bal acc↑)}} & \shortstack[t]{\textbf{target color (bal acc↑)}} & \shortstack[t]{\textbf{target shape (bal acc↑)}} & \shortstack[t]{\textbf{target material (bal acc↑)}} & \shortstack[t]{\textbf{target size (bal acc↑)}} \\
\textit{Baseline} & & \cellcolor{colhighlight}1.00 & \cellcolor{colhighlight}0.33 & 0.12 & 0.33 & 0.50 & 0.50 \\
\hline
\textbf{Qwen3-VL-8B} & hidden state (best layer) & \cellcolor{colhighlight}0.01$_{0.00}$ & \cellcolor{colhighlight}1.00$_{0.00}$ & 0.85$_{0.00}$ & 1.00$_{0.00}$ & 0.84$_{0.00}$ & 0.89$_{0.00}$ \\
 & trajectory-all yes/no & \cellcolor{colhighlight}0.01$_{0.00}$ & \cellcolor{colhighlight}1.00$_{0.00}$ & 0.84$_{0.00}$ & 1.00$_{0.00}$ & 0.79$_{0.00}$ & 0.87$_{0.00}$ \\
 & trajectory-all yes/no (CNN) & \cellcolor{colhighlight}0.03$_{0.00}$ & \cellcolor{colhighlight}1.00$_{0.00}$ & 0.77$_{0.00}$ & 1.00$_{0.00}$ & 0.75$_{0.00}$ & 0.87$_{0.00}$ \\
\rowcolor{rowhighlight}  & trajectory-2 & \cellcolor{cellhighlight}0.03$_{0.00}$ & \cellcolor{cellhighlight}1.00$_{0.00}$ & 0.62$_{0.00}$ & 0.93$_{0.00}$ & 0.69$_{0.00}$ & 0.81$_{0.00}$ \\
\rowcolor{rowhighlight}  & trajectory-2 (CNN) & \cellcolor{cellhighlight}0.03$_{0.00}$ & \cellcolor{cellhighlight}1.00$_{0.00}$ & 0.65$_{0.00}$ & 0.96$_{0.00}$ & 0.73$_{0.00}$ & 0.84$_{0.00}$ \\
 & logits-2 & \cellcolor{colhighlight}0.78$_{0.00}$ & \cellcolor{colhighlight}0.37$_{0.00}$ & 0.14$_{0.00}$ & 0.37$_{0.00}$ & \textcolor{gray}{0.50$_{0.00}$} & 0.56$_{0.00}$ \\
 & logits-all yes/no & \cellcolor{colhighlight}0.10$_{0.00}$ & \cellcolor{colhighlight}0.72$_{0.00}$ & 0.36$_{0.00}$ & 0.54$_{0.00}$ & 0.57$_{0.00}$ & 0.65$_{0.00}$ \\
 & logits-1L & \cellcolor{colhighlight}0.08$_{0.00}$ & \cellcolor{colhighlight}0.90$_{0.00}$ & 0.84$_{0.00}$ & 0.87$_{0.00}$ & 0.62$_{0.00}$ & 0.73$_{0.00}$ \\
\rowcolor{rowhighlight}  & logits-2L & \cellcolor{cellhighlight}0.05$_{0.00}$ & \cellcolor{cellhighlight}0.98$_{0.00}$ & 0.85$_{0.00}$ & 0.98$_{0.00}$ & 0.71$_{0.00}$ & 0.81$_{0.00}$ \\
 & logits-5L & \cellcolor{colhighlight}0.04$_{0.00}$ & \cellcolor{colhighlight}1.00$_{0.00}$ & 0.85$_{0.00}$ & 1.00$_{0.00}$ & 0.80$_{0.00}$ & 0.84$_{0.00}$ \\
 & logits-15L & \cellcolor{colhighlight}0.02$_{0.00}$ & \cellcolor{colhighlight}1.00$_{0.00}$ & 0.85$_{0.00}$ & 1.00$_{0.00}$ & 0.82$_{0.00}$ & 0.87$_{0.00}$ \\
 & imputed-1L & \cellcolor{colhighlight}0.09$_{0.00}$ & \cellcolor{colhighlight}0.74$_{0.00}$ & 0.37$_{0.00}$ & 0.52$_{0.00}$ & 0.56$_{0.00}$ & 0.63$_{0.00}$ \\
 & imputed-2L & \cellcolor{colhighlight}0.08$_{0.00}$ & \cellcolor{colhighlight}0.81$_{0.00}$ & 0.45$_{0.00}$ & 0.59$_{0.00}$ & 0.60$_{0.00}$ & 0.66$_{0.00}$ \\
 & imputed-5L & \cellcolor{colhighlight}0.08$_{0.00}$ & \cellcolor{colhighlight}0.82$_{0.00}$ & 0.48$_{0.00}$ & 0.63$_{0.00}$ & 0.61$_{0.00}$ & 0.66$_{0.00}$ \\
 & imputed-15L & \cellcolor{colhighlight}0.08$_{0.00}$ & \cellcolor{colhighlight}0.83$_{0.00}$ & 0.48$_{0.00}$ & 0.64$_{0.00}$ & 0.61$_{0.00}$ & 0.67$_{0.00}$ \\
\hline
\textbf{LLaVA-7B} & hidden state (best layer) & \cellcolor{colhighlight}0.01$_{0.00}$ & \cellcolor{colhighlight}1.00$_{0.00}$ & 0.81$_{0.00}$ & 1.00$_{0.00}$ & 0.80$_{0.00}$ & 0.85$_{0.00}$ \\
 & trajectory-all yes/no & \cellcolor{colhighlight}0.04$_{0.00}$ & \cellcolor{colhighlight}1.00$_{0.00}$ & 0.79$_{0.00}$ & 1.00$_{0.00}$ & 0.74$_{0.00}$ & 0.82$_{0.00}$ \\
 & trajectory-all yes/no (CNN) & \cellcolor{colhighlight}0.04$_{0.00}$ & \cellcolor{colhighlight}1.00$_{0.00}$ & 0.76$_{0.00}$ & 1.00$_{0.00}$ & 0.72$_{0.00}$ & 0.82$_{0.00}$ \\
\rowcolor{rowhighlight}  & trajectory-2 & \cellcolor{cellhighlight}0.08$_{0.00}$ & \cellcolor{cellhighlight}0.95$_{0.00}$ & 0.68$_{0.00}$ & 0.99$_{0.00}$ & 0.63$_{0.00}$ & 0.76$_{0.00}$ \\
\rowcolor{rowhighlight}  & trajectory-2 (CNN) & \cellcolor{cellhighlight}0.07$_{0.00}$ & \cellcolor{cellhighlight}0.82$_{0.00}$ & 0.72$_{0.00}$ & 0.99$_{0.00}$ & 0.68$_{0.00}$ & 0.79$_{0.00}$ \\
 & logits-2 & \cellcolor{colhighlight}0.77$_{0.00}$ & \cellcolor{colhighlight}0.49$_{0.00}$ & 0.14$_{0.00}$ & 0.34$_{0.00}$ & 0.52$_{0.00}$ & 0.55$_{0.00}$ \\
 & logits-all yes/no & \cellcolor{colhighlight}0.39$_{0.00}$ & \cellcolor{colhighlight}0.68$_{0.00}$ & 0.41$_{0.00}$ & 0.49$_{0.00}$ & 0.59$_{0.00}$ & 0.65$_{0.00}$ \\
 & logits-1L & \cellcolor{colhighlight}0.17$_{0.00}$ & \cellcolor{colhighlight}0.83$_{0.00}$ & 0.78$_{0.00}$ & 0.86$_{0.00}$ & 0.67$_{0.00}$ & 0.70$_{0.00}$ \\
\rowcolor{rowhighlight}  & logits-2L & \cellcolor{cellhighlight}0.11$_{0.00}$ & \cellcolor{cellhighlight}0.97$_{0.00}$ & 0.79$_{0.00}$ & 0.96$_{0.00}$ & 0.71$_{0.00}$ & 0.77$_{0.00}$ \\
 & logits-5L & \cellcolor{colhighlight}0.06$_{0.00}$ & \cellcolor{colhighlight}1.00$_{0.00}$ & 0.80$_{0.00}$ & 1.00$_{0.00}$ & 0.74$_{0.00}$ & 0.81$_{0.00}$ \\
 & logits-15L & \cellcolor{colhighlight}0.03$_{0.00}$ & \cellcolor{colhighlight}1.00$_{0.00}$ & 0.81$_{0.00}$ & 1.00$_{0.00}$ & 0.77$_{0.00}$ & 0.83$_{0.00}$ \\
 & imputed-1L & \cellcolor{colhighlight}0.26$_{0.00}$ & \cellcolor{colhighlight}0.77$_{0.00}$ & 0.48$_{0.00}$ & 0.61$_{0.00}$ & 0.58$_{0.00}$ & 0.67$_{0.00}$ \\
 & imputed-2L & \cellcolor{colhighlight}0.21$_{0.00}$ & \cellcolor{colhighlight}0.80$_{0.00}$ & 0.54$_{0.00}$ & 0.71$_{0.00}$ & 0.60$_{0.00}$ & 0.71$_{0.00}$ \\
 & imputed-5L & \cellcolor{colhighlight}0.21$_{0.00}$ & \cellcolor{colhighlight}0.81$_{0.00}$ & 0.56$_{0.00}$ & 0.73$_{0.00}$ & 0.60$_{0.00}$ & 0.71$_{0.00}$ \\
 & imputed-15L & \cellcolor{colhighlight}0.21$_{0.00}$ & \cellcolor{colhighlight}0.81$_{0.00}$ & 0.56$_{0.00}$ & 0.73$_{0.00}$ & 0.60$_{0.00}$ & 0.71$_{0.00}$ \\
\hline
\textbf{Llama-3.2-11B} & hidden state (best layer) & \cellcolor{colhighlight}0.01$_{0.00}$ & \cellcolor{colhighlight}1.00$_{0.00}$ & 0.78$_{0.00}$ & 1.00$_{0.00}$ & 0.78$_{0.00}$ & 0.83$_{0.00}$ \\
 & trajectory-all yes/no & \cellcolor{colhighlight}0.03$_{0.00}$ & \cellcolor{colhighlight}1.00$_{0.00}$ & 0.78$_{0.00}$ & 1.00$_{0.00}$ & 0.75$_{0.00}$ & 0.80$_{0.00}$ \\
 & trajectory-all yes/no (CNN) & \cellcolor{colhighlight}0.04$_{0.00}$ & \cellcolor{colhighlight}1.00$_{0.00}$ & 0.74$_{0.00}$ & 1.00$_{0.00}$ & 0.74$_{0.00}$ & 0.81$_{0.00}$ \\
\rowcolor{rowhighlight}  & trajectory-2 & \cellcolor{cellhighlight}0.07$_{0.00}$ & \cellcolor{cellhighlight}0.98$_{0.00}$ & 0.66$_{0.00}$ & 1.00$_{0.00}$ & 0.66$_{0.00}$ & 0.73$_{0.00}$ \\
\rowcolor{rowhighlight}  & trajectory-2 (CNN) & \cellcolor{cellhighlight}0.05$_{0.00}$ & \cellcolor{cellhighlight}0.99$_{0.00}$ & 0.62$_{0.00}$ & 1.00$_{0.00}$ & 0.70$_{0.00}$ & 0.78$_{0.00}$ \\
 & logits-2 & \cellcolor{colhighlight}0.77$_{0.00}$ & \cellcolor{colhighlight}0.51$_{0.00}$ & 0.14$_{0.00}$ & 0.37$_{0.00}$ & 0.51$_{0.00}$ & 0.55$_{0.00}$ \\
 & logits-all yes/no & \cellcolor{colhighlight}0.21$_{0.00}$ & \cellcolor{colhighlight}0.83$_{0.00}$ & 0.29$_{0.00}$ & 0.53$_{0.00}$ & 0.57$_{0.00}$ & 0.63$_{0.00}$ \\
 & logits-1L & \cellcolor{colhighlight}0.12$_{0.00}$ & \cellcolor{colhighlight}0.94$_{0.00}$ & 0.60$_{0.00}$ & 0.90$_{0.00}$ & 0.60$_{0.00}$ & 0.67$_{0.00}$ \\
\rowcolor{rowhighlight}  & logits-2L & \cellcolor{cellhighlight}0.07$_{0.00}$ & \cellcolor{cellhighlight}0.97$_{0.00}$ & 0.75$_{0.00}$ & 0.98$_{0.00}$ & 0.65$_{0.00}$ & 0.72$_{0.00}$ \\
 & logits-5L & \cellcolor{colhighlight}0.05$_{0.00}$ & \cellcolor{colhighlight}0.99$_{0.00}$ & 0.78$_{0.00}$ & 1.00$_{0.00}$ & 0.72$_{0.00}$ & 0.77$_{0.00}$ \\
 & logits-15L & \cellcolor{colhighlight}0.03$_{0.00}$ & \cellcolor{colhighlight}1.00$_{0.00}$ & 0.78$_{0.00}$ & 1.00$_{0.00}$ & 0.76$_{0.00}$ & 0.81$_{0.00}$ \\
 & imputed-1L & \cellcolor{colhighlight}0.16$_{0.00}$ & \cellcolor{colhighlight}0.86$_{0.00}$ & 0.30$_{0.00}$ & 0.55$_{0.00}$ & 0.58$_{0.00}$ & 0.64$_{0.00}$ \\
 & imputed-2L & \cellcolor{colhighlight}0.15$_{0.00}$ & \cellcolor{colhighlight}0.88$_{0.00}$ & 0.33$_{0.00}$ & 0.59$_{0.00}$ & 0.59$_{0.00}$ & 0.66$_{0.00}$ \\
 & imputed-5L & \cellcolor{colhighlight}0.14$_{0.00}$ & \cellcolor{colhighlight}0.89$_{0.00}$ & 0.36$_{0.00}$ & 0.62$_{0.00}$ & 0.60$_{0.00}$ & 0.66$_{0.00}$ \\
 & imputed-15L & \cellcolor{colhighlight}0.14$_{0.00}$ & \cellcolor{colhighlight}0.88$_{0.00}$ & 0.36$_{0.00}$ & 0.62$_{0.00}$ & 0.60$_{0.00}$ & 0.66$_{0.00}$ \\
\hline
\end{tabular}
}
\caption{Probe performance on CLEVR: response-altering attributes and target object attributes; all probes are linear unless otherwise stated. Image noise are MSE averages over diffusion, glass blur, and motion blur. Subscripts are standard error of the mean. Probe performance within chance is shown in gray. \hlrow{Highlighted rows} for each model indicate representations that have equal dimensionality; noise-related columns are shown in gray.}

\label{tab:clevr_table1}
\end{table}

\begin{table}[ht]
\centering
\resizebox{\textwidth}{!}{%
\begin{tabular}{llcccccc}
\hline
\multicolumn{2}{l}{} & \multicolumn{6}{c}{\textbf{Predicted Target}} \\
\cline{3-8}
\textbf{Model} & \textbf{Representation Type} & \shortstack[t]{\textbf{background \# obj (MAE↓)}} & \shortstack[t]{\textbf{background color (bal acc↑)}} & \shortstack[t]{\textbf{background shape (bal acc↑)}} & \shortstack[t]{\textbf{background material (bal acc↑)}} & \shortstack[t]{\textbf{background size (bal acc↑)}} & \shortstack[t]{\textbf{grid location (bal acc↑)}} \\
\textit{Baseline} & & 2.00 & 0.50 & 0.50 & 0.50 & 0.50 & 0.11 \\
\hline
\textbf{Qwen3-VL-8B} & hidden state (best layer) & 0.60$_{0.00}$ & 0.87$_{0.00}$ & 0.83$_{0.00}$ & 0.67$_{0.01}$ & 0.65$_{0.00}$ & 0.37$_{0.00}$ \\
 & trajectory-all yes/no & 0.74$_{0.00}$ & 0.77$_{0.00}$ & 0.75$_{0.00}$ & 0.57$_{0.00}$ & 0.63$_{0.00}$ & 0.33$_{0.00}$ \\
 & trajectory-all yes/no (CNN) & 0.75$_{0.00}$ & 0.67$_{0.00}$ & 0.76$_{0.00}$ & 0.56$_{0.00}$ & 0.62$_{0.00}$ & 0.21$_{0.00}$ \\
\rowcolor{rowhighlight}  & trajectory-2 & 1.13$_{0.00}$ & 0.63$_{0.00}$ & 0.58$_{0.00}$ & \textcolor{gray}{0.50$_{0.00}$} & \textcolor{gray}{0.50$_{0.00}$} & 0.21$_{0.00}$ \\
\rowcolor{rowhighlight}  & trajectory-2 (CNN) & 0.94$_{0.00}$ & 0.64$_{0.00}$ & 0.71$_{0.00}$ & 0.54$_{0.00}$ & 0.57$_{0.00}$ & 0.20$_{0.00}$ \\
 & logits-2 & 1.78$_{0.01}$ & \textcolor{gray}{0.50$_{0.00}$} & \textcolor{gray}{0.50$_{0.00}$} & \textcolor{gray}{0.50$_{0.00}$} & \textcolor{gray}{0.50$_{0.00}$} & \textcolor{gray}{0.11$_{0.00}$} \\
 & logits-all yes/no & 1.67$_{0.01}$ & 0.52$_{0.00}$ & \textcolor{gray}{0.50$_{0.00}$} & \textcolor{gray}{0.50$_{0.00}$} & \textcolor{gray}{0.50$_{0.00}$} & 0.13$_{0.00}$ \\
 & logits-1L & 1.51$_{0.01}$ & 0.57$_{0.00}$ & 0.56$_{0.00}$ & \textcolor{gray}{0.50$_{0.00}$} & \textcolor{gray}{0.50$_{0.00}$} & 0.18$_{0.00}$ \\
\rowcolor{rowhighlight}  & logits-2L & 1.27$_{0.01}$ & 0.61$_{0.00}$ & 0.64$_{0.00}$ & \textcolor{gray}{0.50$_{0.00}$} & 0.51$_{0.00}$ & 0.25$_{0.00}$ \\
 & logits-5L & 0.99$_{0.00}$ & 0.67$_{0.00}$ & 0.66$_{0.00}$ & 0.51$_{0.00}$ & \textcolor{gray}{0.50$_{0.00}$} & 0.31$_{0.00}$ \\
 & logits-15L & 0.84$_{0.00}$ & 0.75$_{0.00}$ & 0.72$_{0.00}$ & 0.51$_{0.00}$ & \textcolor{gray}{0.50$_{0.00}$} & 0.35$_{0.00}$ \\
 & imputed-1L & 1.67$_{0.01}$ & 0.53$_{0.00}$ & 0.51$_{0.00}$ & \textcolor{gray}{0.50$_{0.00}$} & \textcolor{gray}{0.50$_{0.00}$} & \textcolor{gray}{0.11$_{0.00}$} \\
 & imputed-2L & 1.58$_{0.01}$ & 0.55$_{0.00}$ & 0.52$_{0.00}$ & \textcolor{gray}{0.50$_{0.00}$} & \textcolor{gray}{0.50$_{0.00}$} & 0.14$_{0.00}$ \\
 & imputed-5L & 1.55$_{0.01}$ & 0.56$_{0.00}$ & 0.53$_{0.00}$ & \textcolor{gray}{0.50$_{0.00}$} & \textcolor{gray}{0.50$_{0.00}$} & 0.15$_{0.00}$ \\
 & imputed-15L & 1.55$_{0.01}$ & 0.56$_{0.00}$ & 0.53$_{0.00}$ & \textcolor{gray}{0.50$_{0.00}$} & \textcolor{gray}{0.50$_{0.00}$} & 0.16$_{0.00}$ \\
\hline
\textbf{LLaVA-7B} & hidden state (best layer) & 0.66$_{0.00}$ & 0.87$_{0.00}$ & 0.79$_{0.00}$ & 0.57$_{0.00}$ & 0.65$_{0.00}$ & 0.30$_{0.00}$ \\
 & trajectory-all yes/no & 0.86$_{0.00}$ & 0.76$_{0.00}$ & 0.76$_{0.00}$ & 0.54$_{0.00}$ & 0.61$_{0.00}$ & 0.27$_{0.00}$ \\
 & trajectory-all yes/no (CNN) & 0.85$_{0.00}$ & 0.70$_{0.00}$ & 0.71$_{0.00}$ & 0.52$_{0.00}$ & 0.55$_{0.00}$ & 0.18$_{0.00}$ \\
\rowcolor{rowhighlight}  & trajectory-2 & 1.19$_{0.01}$ & 0.63$_{0.00}$ & 0.69$_{0.00}$ & \textcolor{gray}{0.50$_{0.00}$} & \textcolor{gray}{0.50$_{0.00}$} & 0.20$_{0.00}$ \\
\rowcolor{rowhighlight}  & trajectory-2 (CNN) & 1.11$_{0.01}$ & 0.64$_{0.00}$ & 0.71$_{0.00}$ & 0.51$_{0.00}$ & 0.51$_{0.00}$ & 0.17$_{0.00}$ \\
 & logits-2 & 1.78$_{0.01}$ & \textcolor{gray}{0.50$_{0.00}$} & \textcolor{gray}{0.50$_{0.00}$} & \textcolor{gray}{0.50$_{0.00}$} & \textcolor{gray}{0.50$_{0.00}$} & \textcolor{gray}{0.11$_{0.00}$} \\
 & logits-all yes/no & 1.63$_{0.01}$ & 0.55$_{0.00}$ & \textcolor{gray}{0.50$_{0.00}$} & \textcolor{gray}{0.50$_{0.00}$} & \textcolor{gray}{0.50$_{0.00}$} & 0.14$_{0.00}$ \\
 & logits-1L & 1.37$_{0.01}$ & 0.66$_{0.00}$ & 0.55$_{0.00}$ & \textcolor{gray}{0.50$_{0.00}$} & \textcolor{gray}{0.50$_{0.00}$} & 0.20$_{0.00}$ \\
\rowcolor{rowhighlight}  & logits-2L & 1.17$_{0.01}$ & 0.70$_{0.00}$ & 0.60$_{0.00}$ & \textcolor{gray}{0.50$_{0.00}$} & \textcolor{gray}{0.50$_{0.00}$} & 0.23$_{0.00}$ \\
 & logits-5L & 1.00$_{0.00}$ & 0.73$_{0.00}$ & 0.67$_{0.00}$ & 0.51$_{0.00}$ & 0.56$_{0.00}$ & 0.26$_{0.00}$ \\
 & logits-15L & 0.87$_{0.00}$ & 0.78$_{0.00}$ & 0.74$_{0.00}$ & \textcolor{gray}{0.50$_{0.00}$} & 0.53$_{0.00}$ & 0.28$_{0.00}$ \\
 & imputed-1L & 1.59$_{0.01}$ & 0.55$_{0.00}$ & 0.51$_{0.00}$ & \textcolor{gray}{0.50$_{0.00}$} & \textcolor{gray}{0.50$_{0.00}$} & 0.17$_{0.00}$ \\
 & imputed-2L & 1.54$_{0.01}$ & 0.56$_{0.00}$ & 0.52$_{0.00}$ & \textcolor{gray}{0.50$_{0.00}$} & \textcolor{gray}{0.50$_{0.00}$} & 0.18$_{0.00}$ \\
 & imputed-5L & 1.51$_{0.01}$ & 0.57$_{0.00}$ & 0.54$_{0.00}$ & \textcolor{gray}{0.50$_{0.00}$} & \textcolor{gray}{0.50$_{0.00}$} & 0.18$_{0.00}$ \\
 & imputed-15L & 1.50$_{0.01}$ & 0.57$_{0.00}$ & 0.54$_{0.00}$ & \textcolor{gray}{0.50$_{0.00}$} & \textcolor{gray}{0.50$_{0.00}$} & 0.18$_{0.00}$ \\
\hline
\textbf{Llama-3.2-11B} & hidden state (best layer) & 0.73$_{0.00}$ & 0.85$_{0.00}$ & 0.77$_{0.00}$ & 0.56$_{0.00}$ & 0.60$_{0.00}$ & 0.28$_{0.00}$ \\
 & trajectory-all yes/no & 0.91$_{0.00}$ & 0.73$_{0.00}$ & 0.75$_{0.00}$ & 0.51$_{0.00}$ & 0.51$_{0.00}$ & 0.27$_{0.00}$ \\
 & trajectory-all yes/no (CNN) & 0.92$_{0.00}$ & 0.69$_{0.00}$ & 0.73$_{0.00}$ & 0.53$_{0.00}$ & 0.60$_{0.00}$ & 0.18$_{0.00}$ \\
\rowcolor{rowhighlight}  & trajectory-2 & 1.32$_{0.01}$ & 0.62$_{0.00}$ & 0.56$_{0.00}$ & \textcolor{gray}{0.50$_{0.00}$} & \textcolor{gray}{0.50$_{0.00}$} & 0.20$_{0.00}$ \\
\rowcolor{rowhighlight}  & trajectory-2 (CNN) & 1.29$_{0.01}$ & 0.64$_{0.00}$ & 0.71$_{0.00}$ & 0.51$_{0.00}$ & 0.56$_{0.00}$ & 0.17$_{0.00}$ \\
 & logits-2 & 1.78$_{0.01}$ & \textcolor{gray}{0.50$_{0.00}$} & \textcolor{gray}{0.50$_{0.00}$} & \textcolor{gray}{0.50$_{0.00}$} & \textcolor{gray}{0.50$_{0.00}$} & \textcolor{gray}{0.11$_{0.00}$} \\
 & logits-all yes/no & 1.65$_{0.01}$ & 0.55$_{0.00}$ & \textcolor{gray}{0.50$_{0.00}$} & \textcolor{gray}{0.50$_{0.00}$} & \textcolor{gray}{0.50$_{0.00}$} & 0.14$_{0.00}$ \\
 & logits-1L & 1.40$_{0.01}$ & 0.61$_{0.00}$ & 0.51$_{0.00}$ & \textcolor{gray}{0.50$_{0.00}$} & \textcolor{gray}{0.50$_{0.00}$} & 0.19$_{0.00}$ \\
\rowcolor{rowhighlight}  & logits-2L & 1.18$_{0.00}$ & 0.67$_{0.00}$ & 0.59$_{0.00}$ & \textcolor{gray}{0.50$_{0.00}$} & \textcolor{gray}{0.50$_{0.00}$} & 0.22$_{0.00}$ \\
 & logits-5L & 1.01$_{0.00}$ & 0.70$_{0.00}$ & 0.65$_{0.00}$ & \textcolor{gray}{0.50$_{0.00}$} & 0.54$_{0.00}$ & 0.25$_{0.00}$ \\
 & logits-15L & 0.88$_{0.00}$ & 0.77$_{0.00}$ & 0.70$_{0.00}$ & 0.53$_{0.00}$ & 0.59$_{0.00}$ & 0.26$_{0.00}$ \\
 & imputed-1L & 1.62$_{0.01}$ & 0.55$_{0.00}$ & 0.51$_{0.00}$ & \textcolor{gray}{0.50$_{0.00}$} & \textcolor{gray}{0.50$_{0.00}$} & 0.13$_{0.00}$ \\
 & imputed-2L & 1.60$_{0.01}$ & 0.55$_{0.00}$ & 0.51$_{0.00}$ & \textcolor{gray}{0.50$_{0.00}$} & \textcolor{gray}{0.50$_{0.00}$} & 0.13$_{0.00}$ \\
 & imputed-5L & 1.59$_{0.01}$ & 0.56$_{0.00}$ & 0.51$_{0.00}$ & \textcolor{gray}{0.50$_{0.00}$} & \textcolor{gray}{0.50$_{0.00}$} & 0.15$_{0.00}$ \\
 & imputed-15L & 1.59$_{0.01}$ & 0.56$_{0.00}$ & 0.51$_{0.00}$ & \textcolor{gray}{0.50$_{0.00}$} & \textcolor{gray}{0.50$_{0.00}$} & 0.15$_{0.00}$ \\
\hline
\end{tabular}
}
\caption{Probe performance on CLEVR: background object attributes and grid location; all probes are linear unless otherwise stated. Image noise are MSE averages over diffusion, glass blur, and motion blur. Subscripts are standard error of the mean. Probe performance within chance is shown in gray. \hlrow{Highlighted rows} for each model indicate representations that have equal dimensionality.}
\label{tab:clevr_table2}
\end{table}

\begin{table}[ht]
\centering
\resizebox{\textwidth}{!}{%
\begin{tabular}{llccccccc}
\hline
\multicolumn{2}{l}{} & \multicolumn{7}{c}{\textbf{Predicted Target}} \\
\cline{3-9}
\textbf{Model} & \textbf{Representation Type} & \cellcolor{colhighlight}\shortstack[t]{\textbf{image noise (MSE↓)}} & \cellcolor{colhighlight}\shortstack[t]{\textbf{object noise (MSE↓)}} & \cellcolor{colhighlight}\shortstack[t]{\textbf{context noise (MSE↓)}} & \cellcolor{colhighlight}\shortstack[t]{\textbf{noise location (bal acc↑)}} & \shortstack[t]{\textbf{saliency (MSE↓)}} & \shortstack[t]{\textbf{category presence (bal acc↑)}} & \shortstack[t]{\textbf{grid location (bal acc↑)}} \\
\textit{Baseline} & & \cellcolor{colhighlight}1.00 & \cellcolor{colhighlight}1.00 & \cellcolor{colhighlight}1.00 & \cellcolor{colhighlight}0.33 & 1.00 & 0.5 & 0.11 \\
\hline
\textbf{Qwen3-VL-8B} & hidden state (best layer) & \cellcolor{colhighlight}0.02$_{0.00}$ & \cellcolor{colhighlight}0.39$_{0.00}$ & \cellcolor{colhighlight}0.11$_{0.00}$ & \cellcolor{colhighlight}0.97$_{0.00}$ & 0.45$_{0.00}$ & 0.94$_{0.00}$ & 0.50$_{0.00}$ \\
 & trajectory-all yes/no & \cellcolor{colhighlight}0.04$_{0.00}$ & \cellcolor{colhighlight}0.45$_{0.00}$ & \cellcolor{colhighlight}0.17$_{0.00}$ & \cellcolor{colhighlight}0.94$_{0.00}$ & 0.50$_{0.00}$ & 0.91$_{0.00}$ & 0.43$_{0.00}$ \\
 & trajectory-all yes/no (CNN) & \cellcolor{colhighlight}0.03$_{0.00}$ & \cellcolor{colhighlight}0.46$_{0.01}$ & \cellcolor{colhighlight}0.22$_{0.00}$ & \cellcolor{colhighlight}0.94$_{0.00}$ & 0.39$_{0.00}$ & 0.83$_{0.00}$ & 0.29$_{0.00}$ \\
\rowcolor{rowhighlight}  & trajectory-2 & \cellcolor{cellhighlight}0.09$_{0.00}$ & \cellcolor{cellhighlight}0.58$_{0.00}$ & \cellcolor{cellhighlight}0.32$_{0.00}$ & \cellcolor{cellhighlight}0.86$_{0.00}$ & 0.58$_{0.00}$ & 0.75$_{0.00}$ & 0.24$_{0.00}$ \\
\rowcolor{rowhighlight}  & trajectory-2 (CNN) & \cellcolor{cellhighlight}0.05$_{0.00}$ & \cellcolor{cellhighlight}0.51$_{0.01}$ & \cellcolor{cellhighlight}0.25$_{0.00}$ & \cellcolor{cellhighlight}0.93$_{0.00}$ & 0.42$_{0.00}$ & 0.90$_{0.00}$ & 0.41$_{0.00}$ \\
 & logits-2 & \cellcolor{colhighlight}0.66$_{0.00}$ & \cellcolor{colhighlight}0.60$_{0.00}$ & \cellcolor{colhighlight}0.67$_{0.00}$ & \cellcolor{colhighlight}0.59$_{0.00}$ & 0.75$_{0.00}$ & \textcolor{gray}{0.50$_{0.00}$} & \textcolor{gray}{0.11$_{0.00}$} \\
 & logits-all yes/no & \cellcolor{colhighlight}0.31$_{0.00}$ & \cellcolor{colhighlight}0.71$_{0.00}$ & \cellcolor{colhighlight}0.78$_{0.01}$ & \cellcolor{colhighlight}0.72$_{0.00}$ & 0.69$_{0.00}$ & \textcolor{gray}{0.50$_{0.00}$} & 0.12$_{0.00}$ \\
\rowcolor{rowhighlight}  & logits-2L & \cellcolor{cellhighlight}0.18$_{0.00}$ & \cellcolor{cellhighlight}0.51$_{0.00}$ & \cellcolor{cellhighlight}0.43$_{0.00}$ & \cellcolor{cellhighlight}0.80$_{0.00}$ & 0.60$_{0.00}$ & 0.73$_{0.00}$ & 0.23$_{0.00}$ \\
 & logits-13L & \cellcolor{colhighlight}0.07$_{0.00}$ & \cellcolor{colhighlight}0.41$_{0.00}$ & \cellcolor{colhighlight}0.24$_{0.00}$ & \cellcolor{colhighlight}0.91$_{0.00}$ & 0.50$_{0.00}$ & 0.92$_{0.00}$ & 0.45$_{0.00}$ \\
 & imputed-2L & \cellcolor{colhighlight}0.29$_{0.00}$ & \cellcolor{colhighlight}0.57$_{0.00}$ & \cellcolor{colhighlight}0.61$_{0.00}$ & \cellcolor{colhighlight}0.71$_{0.00}$ & 0.71$_{0.00}$ & 0.52$_{0.00}$ & 0.12$_{0.00}$ \\
 & imputed-13L & \cellcolor{colhighlight}0.23$_{0.00}$ & \cellcolor{colhighlight}0.55$_{0.00}$ & \cellcolor{colhighlight}0.55$_{0.00}$ & \cellcolor{colhighlight}0.75$_{0.00}$ & 0.65$_{0.00}$ & 0.58$_{0.00}$ & 0.15$_{0.00}$ \\
\hline
\textbf{LLaVA-7B} & hidden state (best layer) & \cellcolor{colhighlight}0.03$_{0.00}$ & \cellcolor{colhighlight}0.40$_{0.00}$ & \cellcolor{colhighlight}0.10$_{0.00}$ & \cellcolor{colhighlight}0.96$_{0.00}$ & 0.47$_{0.00}$ & 0.94$_{0.00}$ & 0.46$_{0.00}$ \\
 & trajectory-all yes/no & \cellcolor{colhighlight}0.09$_{0.00}$ & \cellcolor{colhighlight}0.52$_{0.00}$ & \cellcolor{colhighlight}0.20$_{0.00}$ & \cellcolor{colhighlight}0.89$_{0.00}$ & 0.52$_{0.00}$ & 0.90$_{0.00}$ & 0.40$_{0.00}$ \\
 & trajectory-all yes/no (CNN) & \cellcolor{colhighlight}0.09$_{0.00}$ & \cellcolor{colhighlight}0.52$_{0.01}$ & \cellcolor{colhighlight}0.18$_{0.00}$ & \cellcolor{colhighlight}0.90$_{0.00}$ & 0.43$_{0.00}$ & 0.84$_{0.00}$ & 0.30$_{0.00}$ \\
\rowcolor{rowhighlight}  & trajectory-2 & \cellcolor{cellhighlight}0.20$_{0.00}$ & \cellcolor{cellhighlight}0.66$_{0.00}$ & \cellcolor{cellhighlight}0.38$_{0.00}$ & \cellcolor{cellhighlight}0.78$_{0.00}$ & 0.60$_{0.00}$ & 0.71$_{0.00}$ & 0.26$_{0.00}$ \\
\rowcolor{rowhighlight}  & trajectory-2 (CNN) & \cellcolor{cellhighlight}0.14$_{0.00}$ & \cellcolor{cellhighlight}0.60$_{0.01}$ & \cellcolor{cellhighlight}0.26$_{0.00}$ & \cellcolor{cellhighlight}0.85$_{0.00}$ & 0.46$_{0.00}$ & 0.93$_{0.00}$ & 0.40$_{0.00}$ \\
 & logits-2 & \cellcolor{colhighlight}0.89$_{0.00}$ & \cellcolor{colhighlight}0.82$_{0.01}$ & \cellcolor{colhighlight}\textcolor{gray}{0.99$_{0.01}$} & \cellcolor{colhighlight}0.52$_{0.00}$ & 0.76$_{0.00}$ & \textcolor{gray}{0.50$_{0.00}$} & \textcolor{gray}{0.11$_{0.00}$} \\
 & logits-all yes/no & \cellcolor{colhighlight}0.67$_{0.00}$ & \cellcolor{colhighlight}0.78$_{0.01}$ & \cellcolor{colhighlight}0.89$_{0.01}$ & \cellcolor{colhighlight}0.56$_{0.00}$ & 0.69$_{0.00}$ & \textcolor{gray}{0.50$_{0.00}$} & \textcolor{gray}{0.11$_{0.00}$} \\
\rowcolor{rowhighlight}  & logits-2L & \cellcolor{cellhighlight}0.27$_{0.00}$ & \cellcolor{cellhighlight}0.67$_{0.00}$ & \cellcolor{cellhighlight}0.54$_{0.00}$ & \cellcolor{cellhighlight}0.70$_{0.00}$ & 0.63$_{0.00}$ & 0.84$_{0.00}$ & 0.19$_{0.00}$ \\
 & logits-10L & \cellcolor{colhighlight}0.13$_{0.00}$ & \cellcolor{colhighlight}0.55$_{0.00}$ & \cellcolor{colhighlight}0.29$_{0.00}$ & \cellcolor{colhighlight}0.85$_{0.00}$ & 0.53$_{0.00}$ & 0.92$_{0.00}$ & 0.37$_{0.00}$ \\
 & imputed-2L & \cellcolor{colhighlight}0.55$_{0.00}$ & \cellcolor{colhighlight}0.76$_{0.00}$ & \cellcolor{colhighlight}0.87$_{0.01}$ & \cellcolor{colhighlight}0.63$_{0.00}$ & 0.72$_{0.00}$ & 0.53$_{0.00}$ & 0.13$_{0.00}$ \\
 & imputed-10L & \cellcolor{colhighlight}0.36$_{0.00}$ & \cellcolor{colhighlight}0.74$_{0.00}$ & \cellcolor{colhighlight}0.71$_{0.00}$ & \cellcolor{colhighlight}0.69$_{0.00}$ & 0.64$_{0.00}$ & 0.62$_{0.00}$ & 0.17$_{0.00}$ \\
\hline
\textbf{Llama-3.2-11B} & hidden state (best layer) & \cellcolor{colhighlight}0.03$_{0.00}$ & \cellcolor{colhighlight}0.42$_{0.00}$ & \cellcolor{colhighlight}0.13$_{0.00}$ & \cellcolor{colhighlight}0.96$_{0.00}$ & 0.48$_{0.00}$ & 0.93$_{0.00}$ & 0.44$_{0.00}$ \\
 & trajectory-all yes/no & \cellcolor{colhighlight}0.05$_{0.00}$ & \cellcolor{colhighlight}0.47$_{0.00}$ & \cellcolor{colhighlight}0.24$_{0.00}$ & \cellcolor{colhighlight}0.92$_{0.00}$ & 0.52$_{0.00}$ & 0.89$_{0.00}$ & 0.38$_{0.00}$ \\
 & trajectory-all yes/no (CNN) & \cellcolor{colhighlight}0.04$_{0.00}$ & \cellcolor{colhighlight}0.49$_{0.01}$ & \cellcolor{colhighlight}0.23$_{0.00}$ & \cellcolor{colhighlight}0.94$_{0.00}$ & 0.42$_{0.00}$ & 0.84$_{0.00}$ & 0.30$_{0.00}$ \\
\rowcolor{rowhighlight}  & trajectory-2 & \cellcolor{cellhighlight}0.13$_{0.00}$ & \cellcolor{cellhighlight}0.63$_{0.00}$ & \cellcolor{cellhighlight}0.42$_{0.00}$ & \cellcolor{cellhighlight}0.82$_{0.00}$ & 0.61$_{0.00}$ & 0.69$_{0.00}$ & 0.24$_{0.00}$ \\
\rowcolor{rowhighlight}  & trajectory-2 (CNN) & \cellcolor{cellhighlight}0.07$_{0.00}$ & \cellcolor{cellhighlight}0.53$_{0.01}$ & \cellcolor{cellhighlight}0.31$_{0.01}$ & \cellcolor{cellhighlight}0.90$_{0.00}$ & 0.47$_{0.00}$ & 0.90$_{0.00}$ & 0.35$_{0.00}$ \\
 & logits-2 & \cellcolor{colhighlight}0.62$_{0.00}$ & \cellcolor{colhighlight}0.63$_{0.00}$ & \cellcolor{colhighlight}0.66$_{0.00}$ & \cellcolor{colhighlight}0.49$_{0.00}$ & 0.80$_{0.00}$ & \textcolor{gray}{0.50$_{0.00}$} & \textcolor{gray}{0.11$_{0.00}$} \\
 & logits-all yes/no & \cellcolor{colhighlight}0.28$_{0.00}$ & \cellcolor{colhighlight}0.74$_{0.00}$ & \cellcolor{colhighlight}0.70$_{0.00}$ & \cellcolor{colhighlight}0.66$_{0.00}$ & 0.72$_{0.00}$ & \textcolor{gray}{0.51$_{0.00}$} & \textcolor{gray}{0.11$_{0.00}$} \\
\rowcolor{rowhighlight}  & logits-2L & \cellcolor{cellhighlight}0.14$_{0.00}$ & \cellcolor{cellhighlight}0.52$_{0.00}$ & \cellcolor{cellhighlight}0.41$_{0.00}$ & \cellcolor{cellhighlight}0.82$_{0.00}$ & 0.61$_{0.00}$ & 0.76$_{0.00}$ & 0.21$_{0.00}$ \\
 & logits-13L & \cellcolor{colhighlight}0.07$_{0.00}$ & \cellcolor{colhighlight}0.40$_{0.00}$ & \cellcolor{colhighlight}0.25$_{0.00}$ & \cellcolor{colhighlight}0.91$_{0.00}$ & 0.52$_{0.00}$ & 0.90$_{0.00}$ & 0.37$_{0.00}$ \\
 & imputed-2L & \cellcolor{colhighlight}0.20$_{0.00}$ & \cellcolor{colhighlight}0.56$_{0.00}$ & \cellcolor{colhighlight}0.52$_{0.00}$ & \cellcolor{colhighlight}0.74$_{0.00}$ & 0.68$_{0.00}$ & 0.54$_{0.00}$ & 0.14$_{0.00}$ \\
 & imputed-13L & \cellcolor{colhighlight}0.19$_{0.00}$ & \cellcolor{colhighlight}0.54$_{0.00}$ & \cellcolor{colhighlight}0.49$_{0.00}$ & \cellcolor{colhighlight}0.77$_{0.00}$ & 0.66$_{0.00}$ & 0.58$_{0.00}$ & 0.16$_{0.00}$ \\
\hline
\end{tabular}
}
\caption{Probe performance on MSCOCO; all probes are linear unless otherwise stated. Image/object/context noise are MSE averages over diffusion, glass blur, and motion blur. Subscripts are standard error of the mean. Probe performance within chance is shown in gray. \hlrow{Highlighted rows} for each model indicate representations that have equal dimensionality; noise-related columns are shown in gray.}
\label{tab:mscoco_table}
\end{table}

\finding{The residual stream encodes information about all scene attributes}\label{find:embeddings} The residual stream (specifically, the best-performing hidden state across all layers) encodes information about all aspects of the scene: response-altering attributes, such as noise level or type, and target-related attributes, such as color or shape in the CLEVR dataset (Table~\ref{tab:clevr_table1}) or saliency in MSCOCO (Table~\ref{tab:mscoco_table}), as well as all background attributes of the scene, such as the color or number of background objects in CLEVR (see Tables~\ref{tab:clevr_table2}) or person presence in MSCOCO (Table~\ref{tab:mscoco_table}) are predicted from the hidden states with near-ceiling accuracy (with the exception of the grid position of the target, which is a harder task). Thus, the residual stream serves as an ``oracle'': it retains all information about the scene regardless of whether a particular attribute was queried or not.

\finding{Tuned lens projections retain much of the information from the full stream about both queried and not-queried attributes}\label{find:trajectory} The tuned lens trajectory of the top-2 logits (trajectory-2) retains much of the information about target- and response-altering attributes, as well as information about background objects for CLEVR and person presence for MSCOCO (see Table~\ref{tab:clevr_table1} for CLEVR and Table~\ref{tab:mscoco_table} for MSCOCO), even though they were not queried. This finding mirrors prior work on secret extraction, which shows that information expected to be secret can be extracted from logit trajectories \citep{cywinski2025eliciting,belrose2023eliciting}. 

\begin{figure}[!h]
    \centering
    \includegraphics[width=\textwidth]{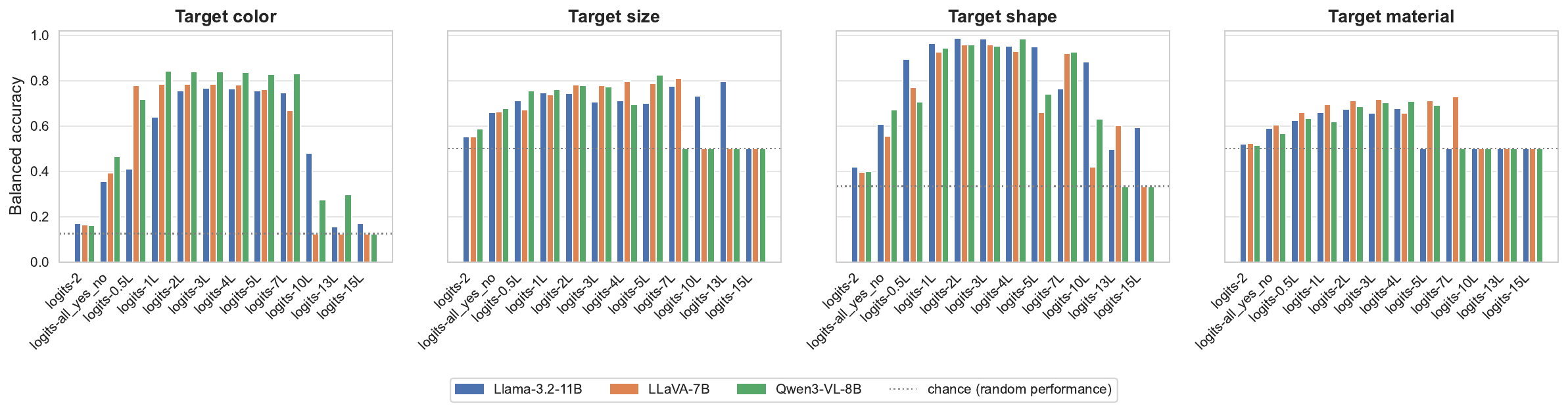}
    \caption{Probe performance on the target attributes in CLEVR as a function of an increasing size of the final logits. $L$ represents the number of layers in the model.}
    \label{fig:target_by_logit_size}
\end{figure}

\finding{Final-layer logits reliably encode information about the target and response-altering attributes}\label{find:logits_relevant} As shown in Tables~\ref{tab:clevr_table1} and ~\ref{tab:mscoco_table}, the information about the response-altering attributes, such as noise level and type, can be predicted with high accuracy from even the top-2 logits (logits-2). For most models, some information about the target attributes can also be recovered from top-2 logits, though probe performance in this case is fairly low. However, as $k$ increases, more target information becomes decodable: target attributes can be predicted reliably above chance starting at the logits-all yes/no level ($k \approx 10$--$13$), with performance peaking at logits-1L ($k \approx 30$--$40$, i.e., roughly the number of model layers; see Figure~\ref{fig:target_by_logit_size}). This suggests that most target properties can be recovered from a relatively small set of top-$k$ final logits. For more details on probe performance as a function of $k$, see Appendix~\ref{app:probe_perf_k_size}.

\begin{figure}[t]
    \centering
    \includegraphics[width=\linewidth]{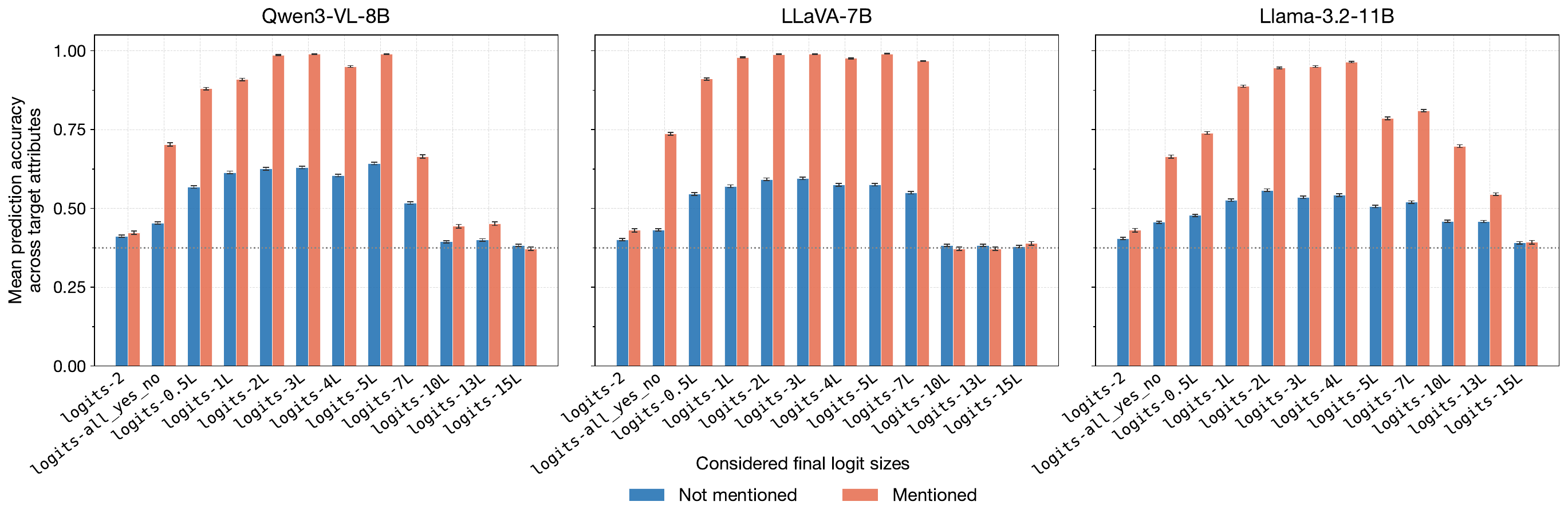}
    \caption{Prediction accuracy for target attributes (averaged across color, material, size) in the CLEVR dataset split by mentioned/not mentioned in the prompt. $L$ represents the number of layers in the model. Bars are averages and error bars are 95\% confidence intervals. Target shape is not included since shape is always mentioned. See Appendix \ref{app:unaggregated_probe_perf} for predictions for each of the attributes without aggregation (Figure~\ref{fig:all_target_by_mentioned}).}
    \label{fig:avg_target_by_mentioned}
\end{figure}

\finding{Final-layer logits contain information about non-queried target attributes}\label{find:logits_not_mentioned} In Finding~\ref{find:logits_relevant}, we showed that the top logits capture information about target attributes such as color or shape. One possibility is that the model is accurately predicting the information that is used to identify the target in the prompt (e.g., color for the prompt ``Is there a blue cylinder in the image?'') as a way to keep track of the target. Another possibility is that other target-related information is leaked into the logits (e.g., target size for the prompt ``Is there a blue cylinder in the image?''). We find that both these possibilities are true: the target attributes regardless of whether they are queried or not can be predicted above chance from all\_yes/no logits and with high accuracy from logits-0.5L (1/2 the number of layers in the model, Figure~\ref{fig:avg_target_by_mentioned}). This has important implications for AI safety. While it is perhaps not surprising that logit trajectories contain information about non-queried attributes of the target (Finding~\ref{find:trajectory}), this information is not typically available to end-users. However, the top few logits are often available, either directly or via sampling. We note that the queried target attributes are overall predicted with higher accuracy (Figure~\ref{fig:avg_target_by_mentioned}), indicating that the top logits are preferentially retaining task-relevant attributes. %

\finding{Final-layer logits contain information about non-target attributes}
\label{find:finding3}
In Table~\ref{tab:clevr_table2}, we see that the top final-layer logits can, in many cases, predict non-target attributes of the scene (such as the number or color of background objects) significantly above chance. While the top-2 logits capture little  information, as $k$ increases we see performance comparable with the top-2 trajectory. This underscores the corollary of Finding~\ref{find:logits_not_mentioned}: access to a relatively modest number of logits allows access to the background information in the scene.

\finding{When controlling for dimensionality, the top-$k$ final-layer logits contain similar information to the logit trajectory} As we saw in Finding~\ref{find:trajectory}, the logit trajectory of the top-2 logits\footnote{As a sanity check, we inspect these trajectories directly in Appendix~\ref{app:dynamics}, replicating the type of analysis typically conducted with tuned/logit lens approaches \citep{belrose2023eliciting,halawi2024overthinkingtruthunderstandinglanguage, cywinski2025eliciting}, and confirm that the evolution of $P(\text{yes})$ across layers responds systematically to both input perturbation and target properties.} has an intermediate predictive power between the best-layer hidden states and top-2 logits, retaining much of the information from the residual stream even about background attributes. What is perhaps surprising is that we see \textit{very similar} information in the top-2L final-layer logits (see highlighted columns in Tables~\ref{tab:clevr_table1},  \ref{tab:clevr_table2} and \ref{tab:mscoco_table}), where $L$ is the number of model layers (between 33 and 41). These 2L last-layer logits exactly match the dimensionality of the top-2 logit trajectory. This is a concerning observation: it suggests that the top last layer logits are equally likely to leak information that we might reasonably expect to be hidden, while being much more easily accessible to the end user. 

\finding{Probes mimicking realistic APIs reliably recover many attributes} Many frontier model APIs (e.g., Google Gemini and several OpenAI models) expose top-20 logits, and serving a model through vllm, a popular tool for model deployment, exposes any number of logits. We simulated this scenario with a masked dataset where only top-20 logits are visible and all other logit values are imputed to that of the smallest visible logit. This setup lets us evaluate attribute recoverability under exactly the information an adversary could observe through a standard API. We find that, while probe accuracy drops in this scenario, we can still predict many non-queried attributes with above-chance accuracy (see ''imputed'' in Tables ~\ref{tab:clevr_table1}, \ref{tab:clevr_table2}, and \ref{tab:mscoco_table}), suggesting that real-life APIs expose enough logits to recover the leaked information. To assess how many queries are needed to carry out such an attack, we perform an ablation on two CLEVR tasks (see Appendix~\ref{app:ablation_n_queries}). Specifically, we vary the number of queries used to construct the training set by randomly subsampling our total dataset. We find that performance starts to saturate around 5000 data points (i.e., queries). We observe above-random performance with as little as 100 data points, even in the case where we only have access to the top-20 logits.

\section{Discussion}
\label{sec:discussion}
An ``optimal'' information bottleneck would compress the information in the input to just the information needed to make a prediction. Our analysis reveals that two natural candidates for such a bottleneck, the top final-layer logits and the logit trajectory obtained using tuned lens, do \textit{not} behave in such a manner. As expected, the information about noise level and type (both factors that appropriately influence the model's response) is already recoverable from the top-2 final-layer logits, indicating that the model retains the response-altering information. However, we also find that larger (but still moderate) sets of top logits retain information about attributes that should not influence the decision, and do not in practice change the probability of answering ``yes" or ``no".

This leakage has direct implications across several areas of machine learning research. \textbf{Safety and privacy}: LLM stochasticity means that leaked information is recoverable via repeated black-box querying, and directly accessible when log-probabilities are exposed. Our findings suggest that, for instance, user-uploaded images could leak sensitive attributes even when queries are unrelated. \textbf{Fairness}: Our MSCOCO and CLEVR attributes are controlled stand-ins for sensitive attributes (race, gender, geography). Future work can test whether such attributes remain decodable when the model abstains from answering, and whether decodability correlates with model bias. \textbf{Distillation}: Our findings offer a mechanistic explanation for distilled models inheriting teacher biases \citep{cloud_owl_traits}. The logit distribution carries non-target attribute information, and matching it transfers those attributes. \textbf{Interpretability}: Rich information persists not only in model internals, i.e., the field's usual focus, but also in the final logits, which have been treated as terminal. \textbf{Calibration and hallucination detection}: Methods like semantic entropy assume logits carry answer-relevant information, but substantial unrequested-attribute information leaks in. This can introduce attribute-correlated bias that aggregate calibration and factuality scores may mask.

\bibliography{main}
\bibliographystyle{plainnat}

\newpage
\appendix
\onecolumn

\section{Statistical analysis of CLEVR dataset}
\subsection{Referring description statistics in CLEVR}
\label{app:clevr_description_length}

Table~\ref{tab:expr_lengths} details the length distribution of the referring expressions in the CLEVR evaluation set. The most common descriptions are two-word expressions (an adjective and a noun, e.g., ``red circle''), comprising 72.8\% of the data. The remainder consists of shorter single-word nouns (11.3\%) and more complex three- to four-word descriptions (15.9\%). Refer to Appendix~\ref{app:flip_rate} for an analysis of model accuracy for each of these description lengths. 

\begin{table}[h]
  \centering
 
  \begin{tabular}{clrrl}
  \toprule
  \textbf{Length} & \textbf{Words} & \textbf{Count} & \textbf{\%} & \textbf{Example} \\
  \midrule
  1 & noun only            & 1{,}758 & 11.3 & \textit{cylinder} \\
  2 & adj + noun           & 11{,}364 & 72.8 & \textit{yellow sphere} \\
  3 & adj + adj + noun     & 1{,}470 &  9.4 & \textit{gray rubber cube} \\
  4 & adj + adj + adj + noun & 1{,}008 &  6.5 & \textit{large green metal cylinder} \\
  \midrule
  \textbf{Total} & & \textbf{15{,}600} & \textbf{100.0} & \\
  \bottomrule
  \end{tabular}
\caption{Distribution of referring expression lengths in the CLEVR evaluation set.}
 \label{tab:expr_lengths}
\end{table}

\subsection{Relationship between model accuracy and description length}
\label{app:flip_rate}

\begin{figure}[!h]
    \centering
    \includegraphics[width=\linewidth]{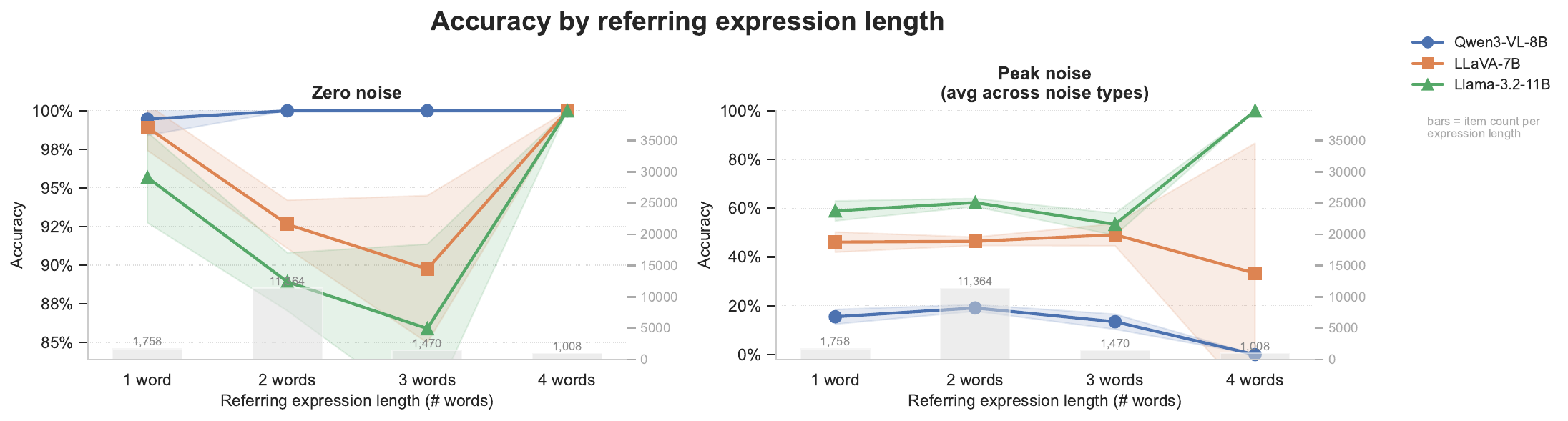}
    \caption{Model accuracy on the CLEVR validation set by referring expression length at zero noise (left) and the highest intensity of noise (right), averaged across the noise types. The gray bars represent the number of samples for each expression length, while the lines show the mean accuracy of each model, with shaded regions denoting 95\% confidence intervals.}
    \label{fig:len_clean_vs_peak}
\end{figure}

Figure~\ref{fig:len_clean_vs_peak} shows the accuracy of the evaluated models' greedily generated responses as a function of referring expression length discussed in Appendix~\ref{app:clevr_description_length}. On clean images, the models generally struggle the most with three-word expressions, with Qwen3-VL-8B demonstrating the strongest baseline performance. However, at the highest noise levels, Qwen3-VL-8B is the most severely affected, achieving the lowest accuracy across all expression lengths. Furthermore, while four-word expressions are generally the most difficult to process under noisy conditions, Llama-3.2-11B presents an increase in accuracy for these longer descriptions. 

We clarify that the accuracy does not merely degrade to a 50\% random baseline in the presence of noise, since the models' systematic shift toward predicting ``No'' on an evaluation set where all images have a ground truth of ``Yes'' ultimately drives the performance down to 0\%. Refer to Appendix~\ref{app:clevr_acc_noise_level_images} for a detailed breakdown of model accuracy by noise type and across different noise levels.

\newpage
\subsection{Impact of noise type and level on model accuracy in CLEVR VQA task}
\label{app:clevr_acc_noise_level_images}

\begin{figure}[!h]
    \centering
    \includegraphics[width=\linewidth]{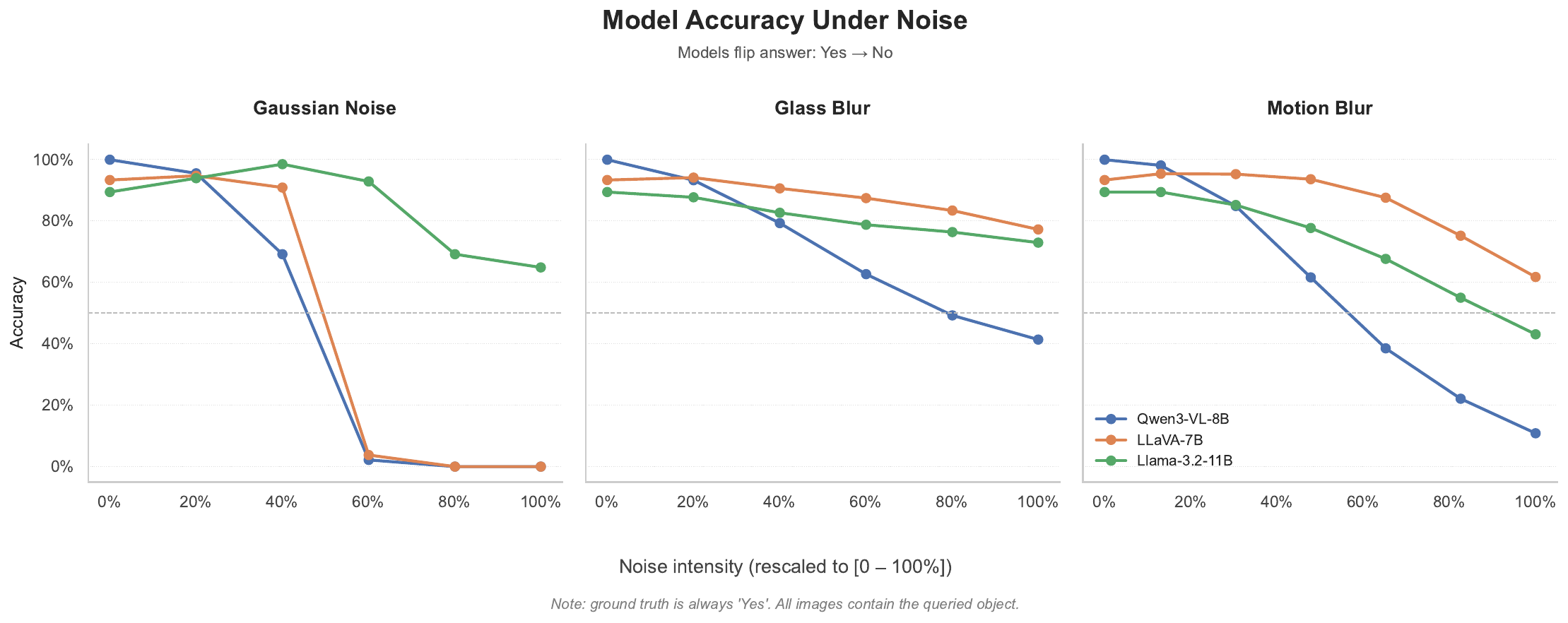}
    \caption{Accuracy degradation across models and noise types on the CLEVR validation set. As noise intensity increases, models increasingly flip their answer from "Yes" to  "No", driving accuracy towards 0.}
    \label{fig:acc_models_per_noise}
\end{figure}

The accuracy of the greedily generated response across models for the three different noise types (Gaussian noise, glass, and motion blur) at varying levels of noise is shown in Figure \ref{fig:acc_models_per_noise}. In the presence of noise, we see that  the models tend to flip their answers from ``Yes'' to ``No''. This behavior leads to significant drops in accuracy, particularly at the highest noise levels. Gaussian noise causes the most severe performance degradation overall. Under this noise type, Qwen3-VL-8B is the most negatively affected, while Llama-3.2-11B proves to be the most robust. Conversely, for the other noise types, LLaVA-7B demonstrates the highest accuracy, followed by Llama-3.2-11B, while Qwen3-VL-8B consistently exhibits the worst performance across the board. 

\begin{figure}[!h]
    \centering
    \includegraphics[width=\linewidth]{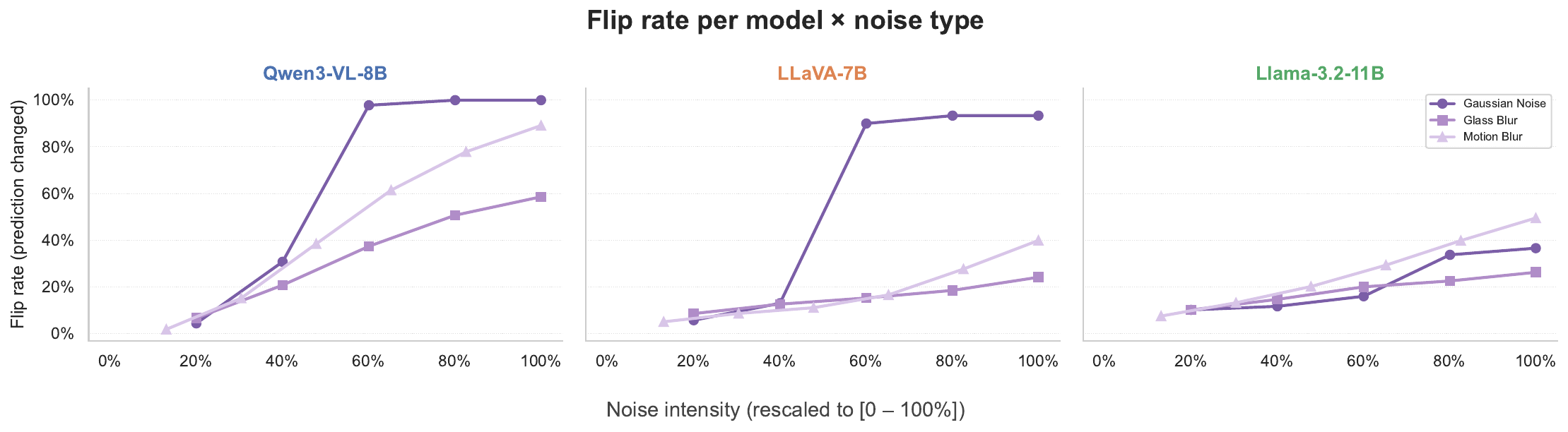}
    \caption{Flip rate analysis across evaluated models, showing the frequency at which models change their initial predictions when exposed to different types and intensities of noise.}
    \label{fig:flip_rate}
\end{figure}

\begin{figure}[!h]
    \centering
    \includegraphics[width=\linewidth]{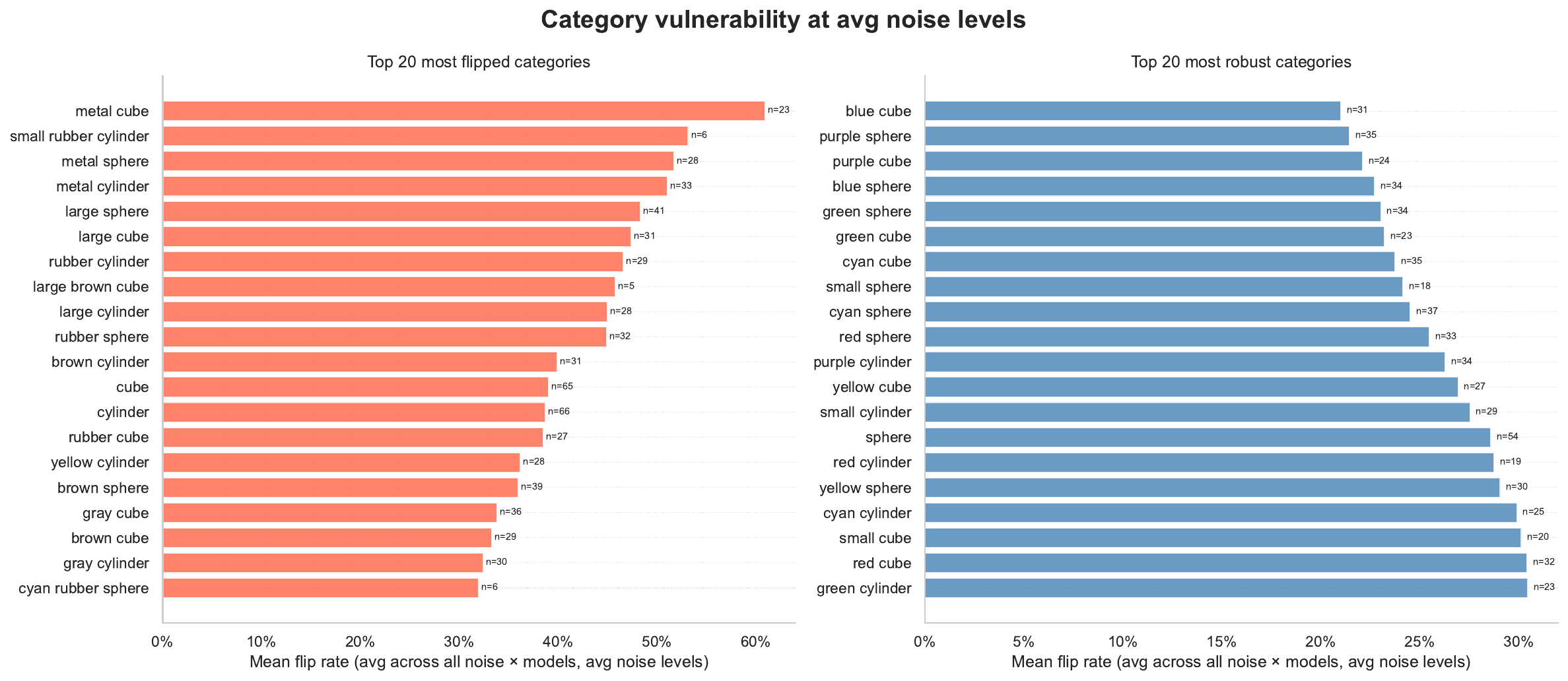}
    \caption{The 20 most vulnerable (left) and 20 most robust (right) categories in CLEVR, ranked by mean flip rate averaged over all noise levels and models. This figure highlights the specific object categories where the models struggle the most under noise, alongside the categories where performance remains robust.}
    \label{fig:cat_vuln}
\end{figure}

Figure~\ref{fig:flip_rate} reports the exact flip rate across varying conditions. Figure~\ref{fig:cat_vuln} provides a fine-grained breakdown of model performance by object type. This highlights the specific categories where the models struggle the most under perturbation, as well as the categories to which they remain the most resilient.

\newpage

\section{Qualitative examples of model question-answering performance on the CLEVR dataset}
\label{app:clevr_acc_noise_level}

\begin{figure}[H]
    \centering
    \includegraphics[width=\linewidth]{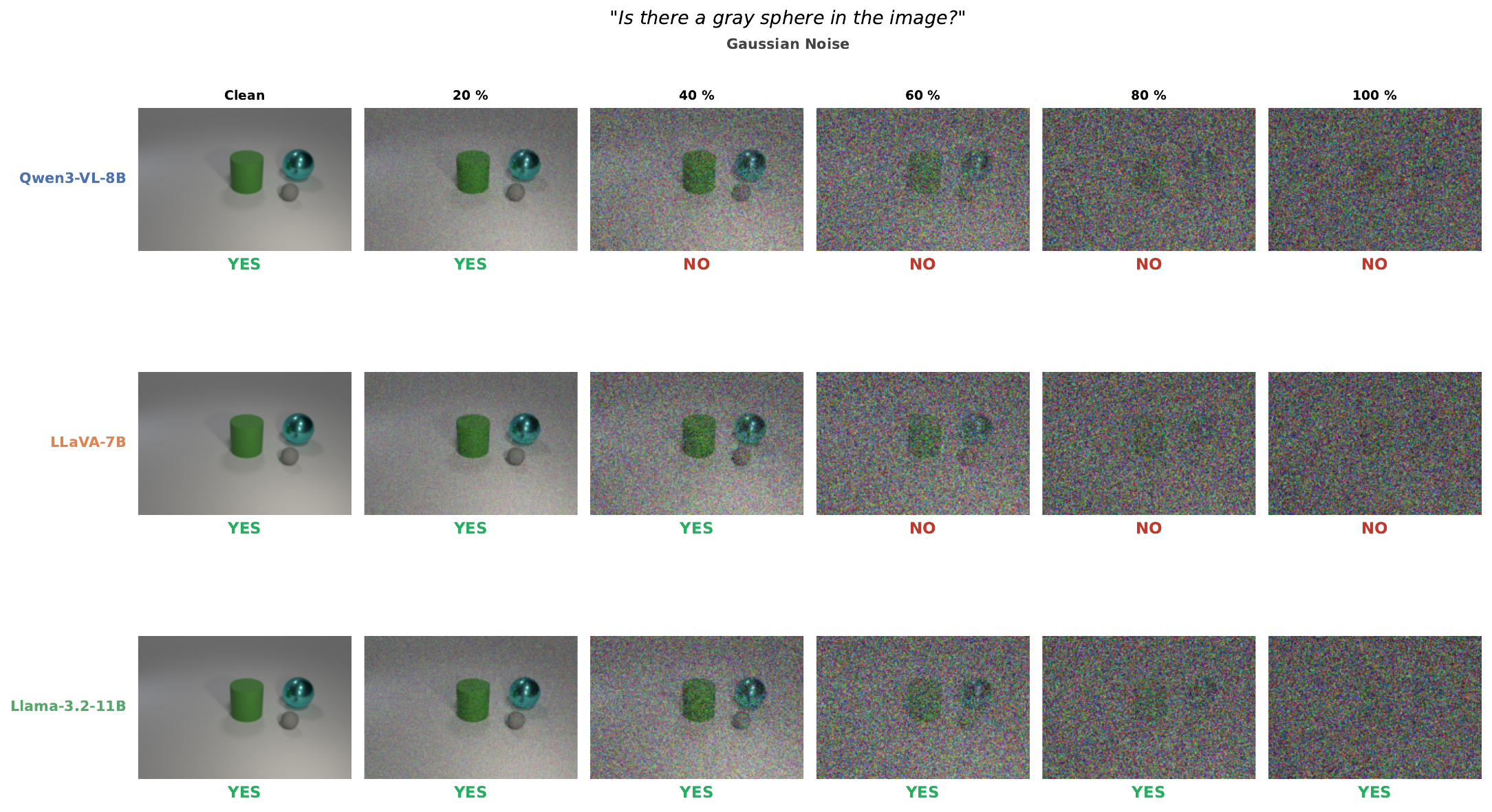}
    \caption{Model predictions under varying intensities of Guassian noise.  Columns display the same CLEVR image at increasing noise intensities (0\% to 100\%); rows denote the evaluated models. The label below each panel indicates the model's predicted answer (``Yes'' or`` No'') to the query: ``Is there a \textlangle target expression\textrangle in the image?''}
    \label{fig:qualitive_example1}
\end{figure}

In this section, we provide qualitative examples of the models' responses on the CLEVR dataset. Figure~\ref{fig:qualitive_example1} shows the models' answers under Gaussian noise at increasing intensity levels. We observe that Qwen3-VL-8B is the most affected by this noise, flipping its prediction to ``No'' even at lower noise levels, whereas Llama-3.2-11B consistently maintains its ``Yes'' prediction across all evaluated levels of Gaussian noise.

\begin{figure}[!h]
    \centering
    \includegraphics[width=\linewidth]{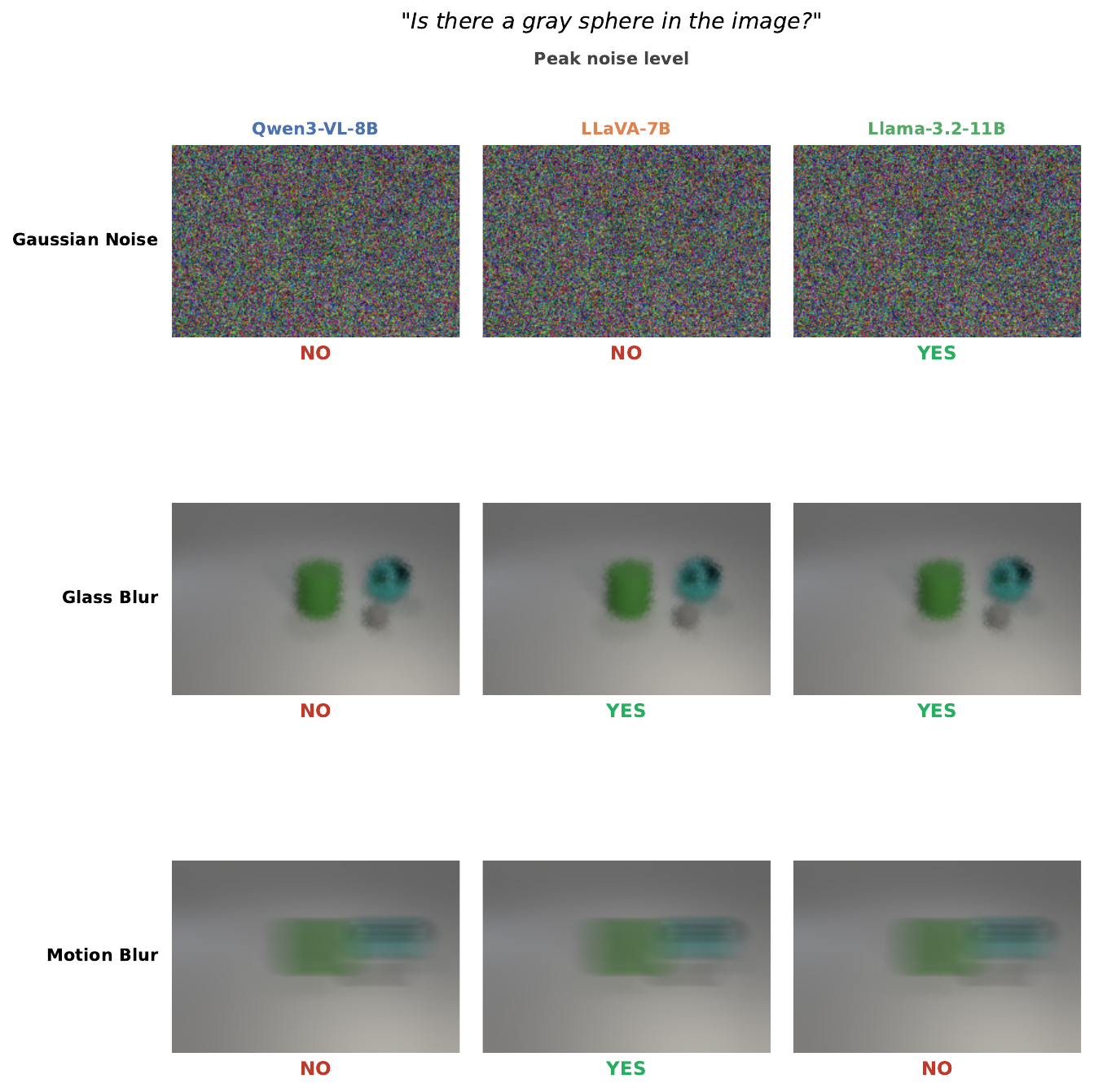}
    \caption{Qualitative examples of model predictions at peak noise intensity across different noise types (rows) for each evaluated model (columns). The label below each panel indicates the model's predicted answer (``Yes'' or ``No'') to the query: ``Is there a [target expression] in the image?''}
    \label{fig:qualitive_example2}
\end{figure}

Figure~\ref{fig:qualitive_example2} presents an example under the highest severity level for the other noise types (glass and motion blur). In line with the quantitative observations in Figure~\ref{fig:acc_models_per_noise}, Qwen3-VL-8B exhibits a strong tendency to flip its answer to ``No'' across different perturbations. Conversely, LlaVA-7B keeps its ``Yes'' answer under glass and motion blur, and Llama-3.2-11B keeps its answer under Gaussian noise and glass blur.

\newpage

\section{Tuned lens: Additional details}\label{app:tuned_lens}

\subsection{Tuned lens training and evaluations}
\label{app:tuned_lens_training}

\begin{figure}[H]
    \centering
    \includegraphics[width=\textwidth]{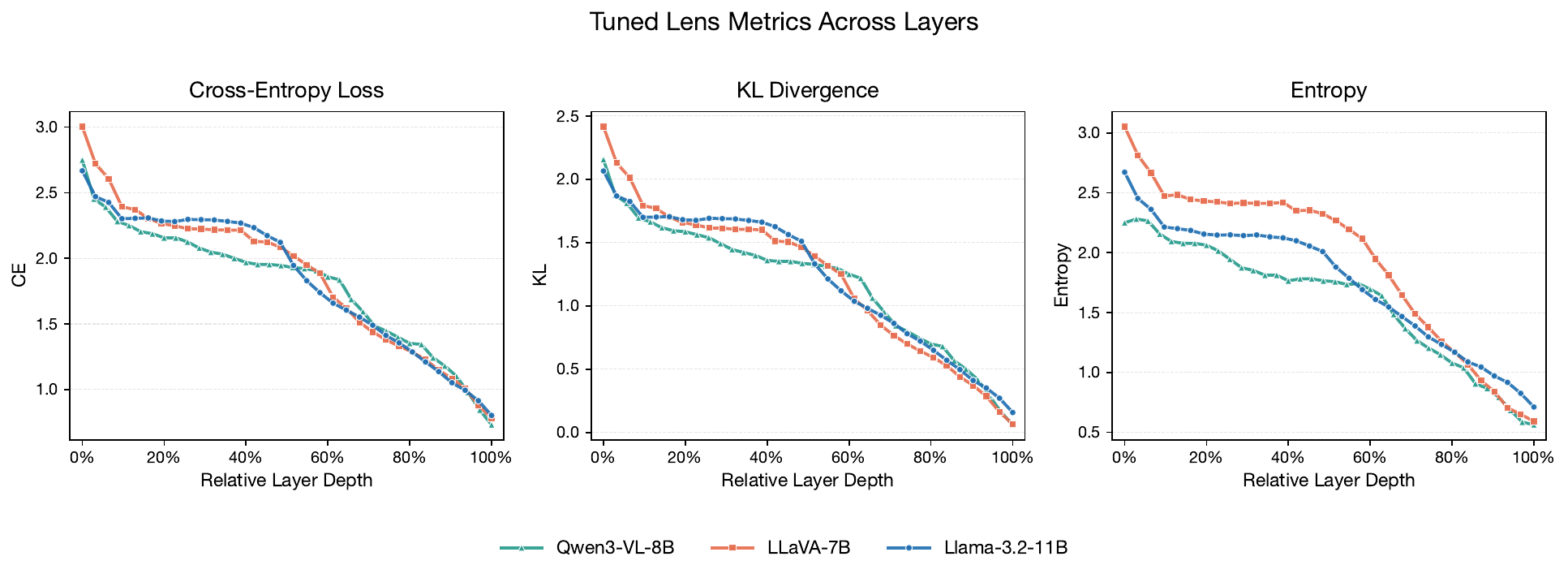}
    \caption{Performance of the trained lenses on the evaluation set. Model layers are normalized to [0, 1] by dividing each layer index by that model's maximum layer index.}
    \label{fig:tuned_lens_metrics}
\end{figure}

We train tuned lenses for all models in our experiments on the validation split of the uncopyrighted subset of the Pile\footnote{https://huggingface.co/datasets/monology/pile-uncopyrighted} following the approach outlined in \cite{belrose2023eliciting} including all the hyper-parameter choices. We evaluate the lenses on a random sample of 16.4M tokens of the test split of the same dataset, again following \cite{belrose2023eliciting}. The performance of the trained lenses on the evaluation set is shown in Figure~\ref{fig:tuned_lens_metrics}.

\subsection{Inspecting the logit trajectories}
\label{app:logit_lens}

\begin{figure}[ht]              
      \centering                                
      \begin{subfigure}[b]{\textwidth}                             
          \centering     
          \includegraphics[width=\textwidth]{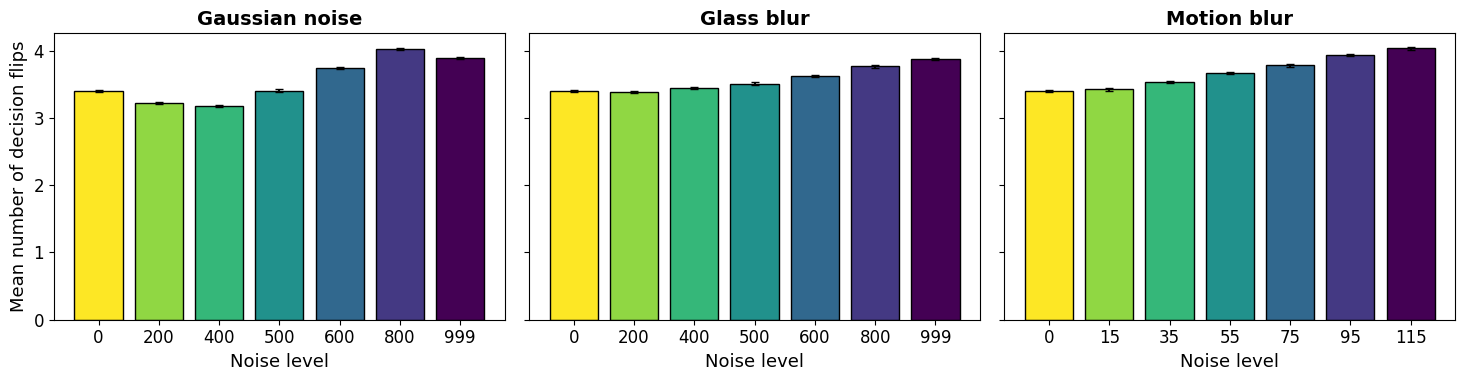}                                                 
          \caption{By noise level}
          \label{fig:noise_decision_volatility}
      \end{subfigure}
      \vspace{0.5em}
      \begin{subfigure}[b]{0.4\textwidth}
          \centering
          \includegraphics[width=\textwidth]{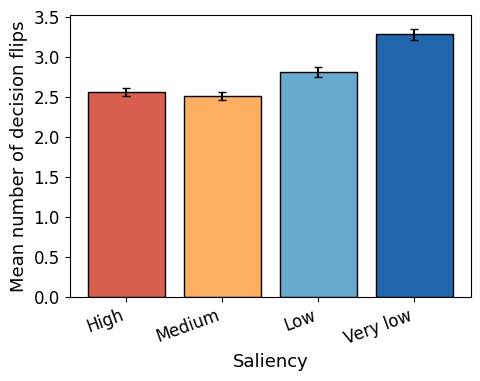}
          \caption{By saliency}
          \label{fig:saliency_decision_volatility}
      \end{subfigure}
      \caption{Mean number of yes/no prediction flips across consecutive layers, pooled across three models under consideration on MS-COCO. Error bars: standard error of the mean. \textbf{(a)}~By noise level:  
  volatility increases with perturbation intensity across all three noise types. \textbf{(b)}~By saliency (unperturbed images only): low saliency objects trigger more decision flips than salient ones. Overall, stronger noise and low saliency seem to prolong internal deliberation.}
      \label{fig:decision_volatility}
\end{figure}

We examine how the models' predictions evolve across layers by inspecting the $P(\text{yes})$ trajectory obtained via the tuned lens on the MSCOCO dataset (see Appendix~\ref{app:mscoco} for details). Figure~\ref{fig:pyes_trajectory} shows these trajectories for three VLMs under varying noise levels~(Panel a) and saliency conditions~(Panel b). In both cases, the models' predictions fluctuate considerably in early and middle layers before converging, and this instability is amplified by stronger noise and lower saliency. To quantify this, Figure~\ref{fig:decision_volatility} counts the number of yes/no flips across consecutive layers. Flip rate increases monotonically with noise intensity~(Panel a) and with decreasing saliency~(Panel b), confirming that both perturbation and low saliency delay the model's convergence to a stable prediction.

\label{app:dynamics}
\begin{figure}[ht]
      \centering
      \begin{subfigure}[b]{\textwidth}
          \centering
          \includegraphics[width=\textwidth]{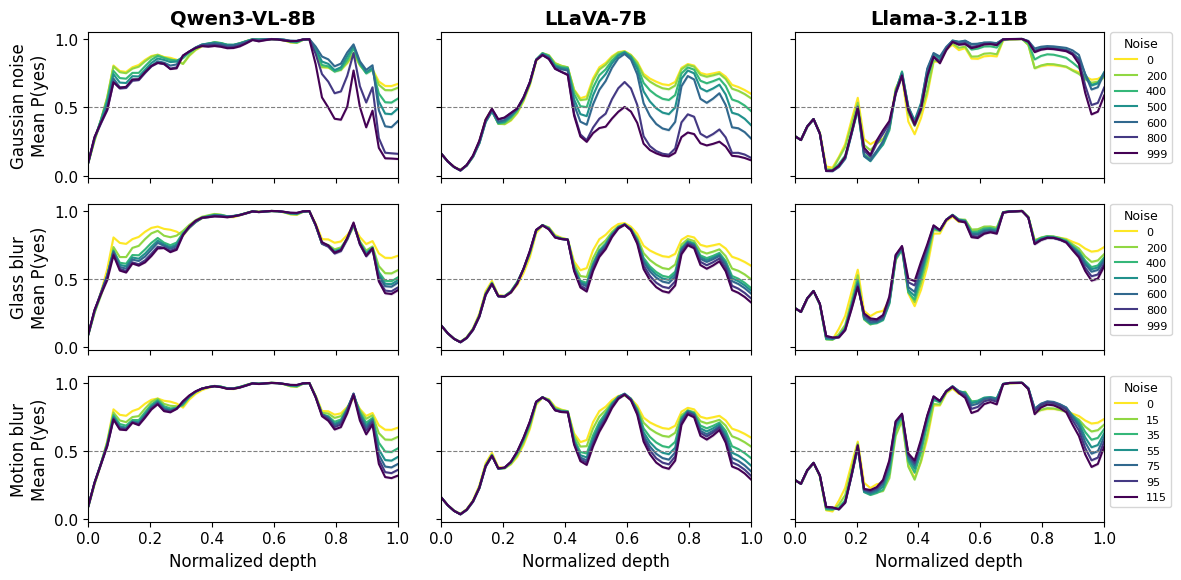}
          \caption{Effect of noise}
          \label{fig:pyes_trajectory_noise}
      \end{subfigure}
      \vspace{0.5em}
      \begin{subfigure}[b]{\textwidth}
          \centering
          \includegraphics[width=\textwidth]{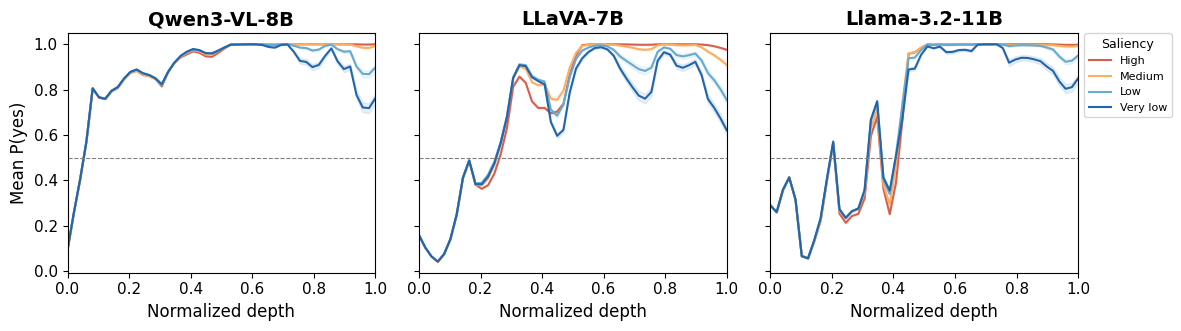}
          \caption{Effect of saliency}
          \label{fig:pyes_trajectory_sal}
      \end{subfigure}
      \caption{Mean $P(\text{yes})$ trajectory across normalized model depth for three models under consideration (columns). \textbf{(a)}~Rows: noise types; color: noise level (yellow = none, purple = maximum). Increasing perturbation suppresses $P(\text{yes})$ toward the final layers, but middle layers often maintain elevated $P(\text{yes})$ even under heavy noise.  \textbf{(b)}~Unperturbed images grouped by saliency. Saliency is binned into four intervals: very low ($<0.1$, bottom ${\sim}20\%$, blue), low ($0.1$--$0.9$, ${\sim}20\%$), medium ($0.9$--$0.99$, ${\sim}30\%$), and high ($\geq 0.99$, top ${\sim}30\%$, red); color runs from blue (very low) to red (high). The distribution is heavily right-skewed because objects explicitly named in the captions receive saliency near 1.0. Lower saliency delays commitment to the final answer and reduces final-layer confidence.}
      \label{fig:pyes_trajectory}
\end{figure}

\clearpage

\section{Choice of representational levels}
\label{app:rep_levels}
\begin{figure}[ht]
    \centering
    \includegraphics[width=\linewidth]{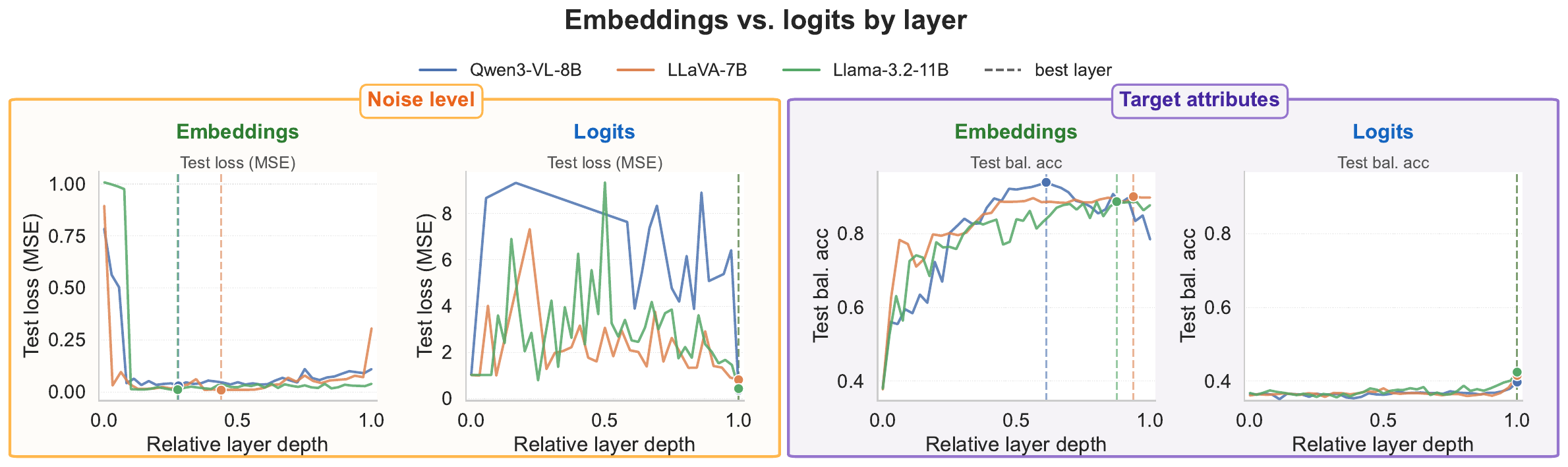}
    \caption{ Test loss for noise-level prediction and balanced accuracy for target attribute classification, shown by normalized layer depth for embeddings (left columns in each pane;) and logits (right columns in each panel). Layer indices are normalized by model depth to allow comparison across architectures. The dashed line marks the
  best-performing layer per model; for embeddings this is consistently intermediate, whereas for logits it is the final layer.}
    \label{fig:emb_logit_1}
\end{figure}
For representing the residual stream, we choose to use the best-performing layer (evaluated on a validation set), as this gives a reasonable "upper bound" on the amount of information available. As we see in Figure~\ref{fig:emb_logit_1}, this typically isn't the last layer; however performance at the last layer is typically comparable with the best layer. %

In principle, we could consider the single-layer tuned-lens projections of the residual stream into logit space as representation levels. However, in practice, we find that the intermediate-layer logits are almost always significantly worse predictors than the final-layer logits (see Figure~\ref{fig:emb_logit_1}).

\section{Probe technical details}
\label{app:probes}

\textbf{Linear probes}. We select the regression model based on the outcome variable. For noise level and saliency, we use ridge regression. For number of other objects, we use Huber regression (epsilon = 1.35, max\_iter = 1000). For all remaining (binary) outcomes, we use logistic regression with L2 regularization and the L-BFGS solver (sklearn defaults). All features are standardized using sklearn's StandardScaler, and regularization strength is selected via cross-validation over a grid.

\textbf{CNN}. For tuned lens trajectories, we additionally use a 1D CNN with three convolutional blocks (16, 32, and 64 filters with kernel sizes 5, 5, and 3, respectively), each followed by ReLU activation and batch normalization, with max pooling after the second and third blocks, followed by two fully connected layers (128 and 32 units) with a dropout of 0.2. We used early stopping with a patience of 50.

\textbf{Training objectives}. For predicting noise levels, we used MSE as our training objective. For predicting number of other objects, we used an L1-like loss (L1 for the CNN; Huber for linear probes due to greater stability). We treated target attribute prediction as a multiclass classification and background object attribute prediction as a multi-label classification, using cross-entropy in both cases.

\section{Probe performance on CLEVR as a function of representation size}
\label{app:probe_perf_k_size}

\subsection{Background attributes}
\label{app:perf_bg_objects}

\begin{figure}[H]
    \centering
    \includegraphics[width=\textwidth]{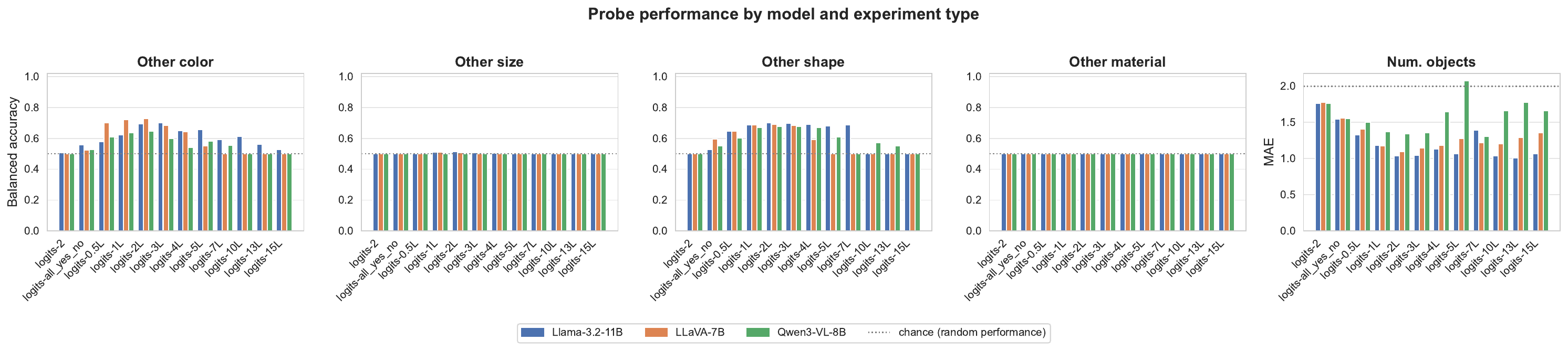}
    \caption{Probe performance on the background attributes as a function of an increasing size of the final logits.}
    \label{fig:other_by_logit_size}
\end{figure}

\subsection{Noise attributes}
\label{app:perf_noise}
\begin{figure}[H]
    \centering
    \includegraphics[width=\textwidth]{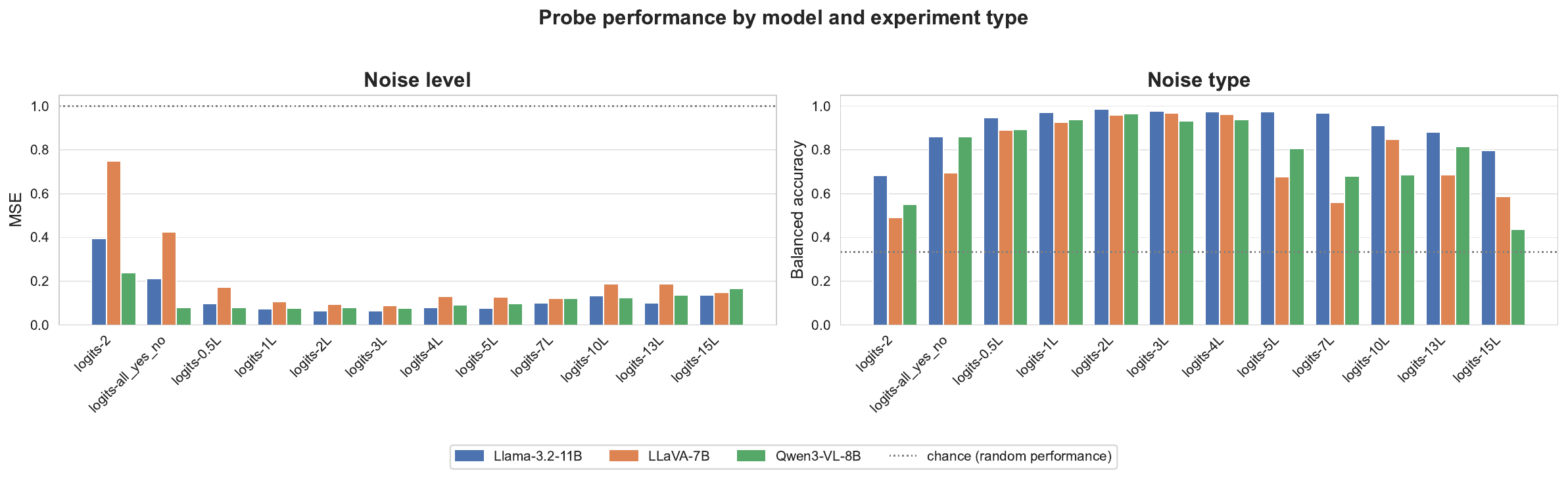}
    \caption{Probe performance on the noise attributes as a function of an increasing size of the final logits.}
    \label{fig:noise_by_logit_size}
\end{figure}

\subsection{Unaggregated target attribute predictions split whether they are queried or not}
\label{app:unaggregated_probe_perf}
\begin{figure}[h]
    \centering
    \includegraphics[width=\linewidth]{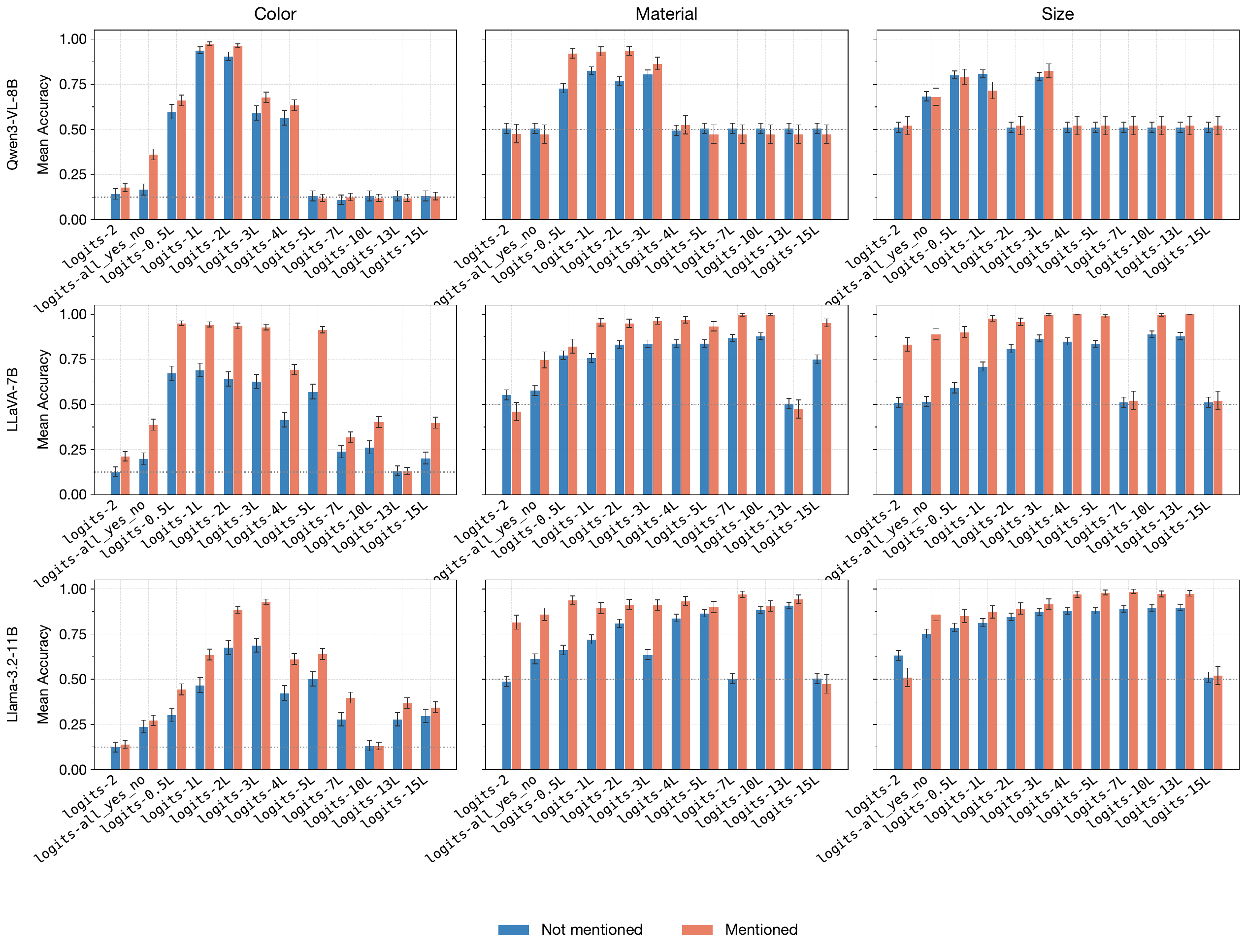}
    \caption{Probe performance on unaggregated target attributes in the CLEVR dataset split by mentioned/not mentioned in the prompt. Bars are averages and error bars are 95\% confidence intervals. Target shape is not included since shape is always mentioned.}
    \label{fig:all_target_by_mentioned}
\end{figure}

In Figure~\ref{fig:all_target_by_mentioned}, we show the per-attribute probe performance on target attributes, split by whether they are mentioned in the prompt. The average of these plots is shown in Figure~\ref{fig:avg_target_by_mentioned}.

\clearpage

\section{Additional information on MSCOCO}
\label{app:mscoco}
\subsection{Illustration of noise perturbations}
\label{app:mscoco_noise}

\begin{figure}[!h] 
      \centering
      \setlength{\tabcolsep}{2pt}                                                                                   
      \renewcommand{\arraystretch}{0.8}
      \begin{tabular}{cccc}                                                                                         
          \includegraphics[width=0.23\textwidth]{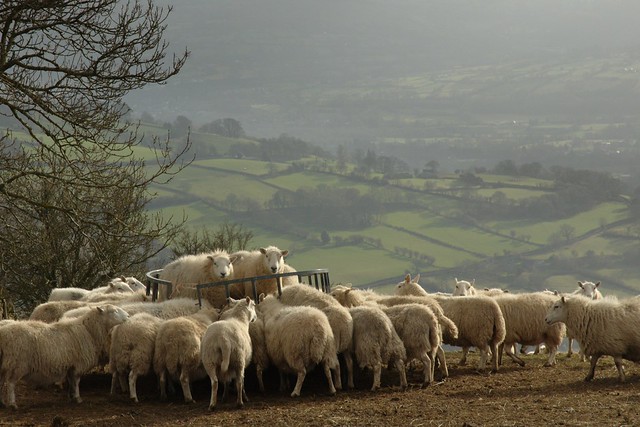} &
          \includegraphics[width=0.23\textwidth]{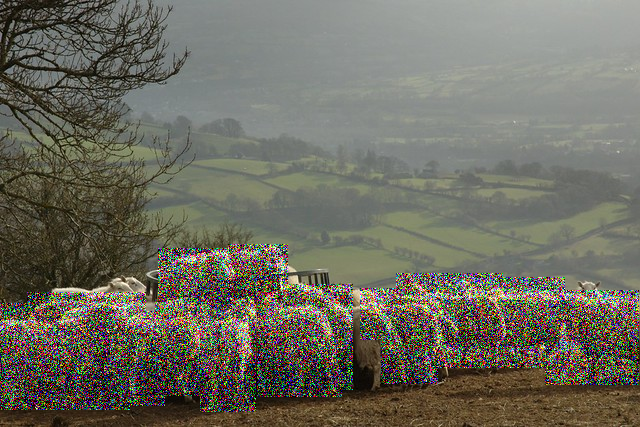} &
          \includegraphics[width=0.23\textwidth]{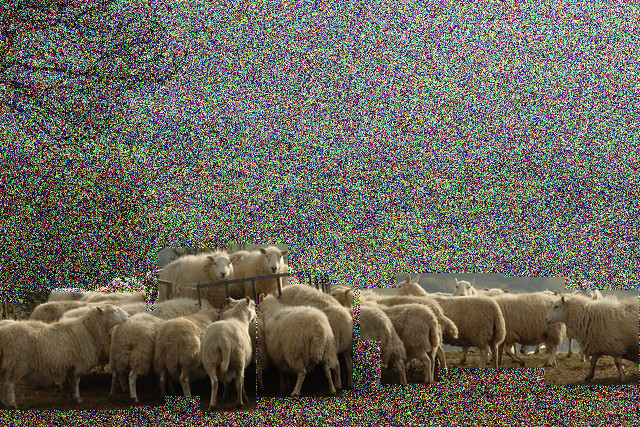} &
          \includegraphics[width=0.23\textwidth]{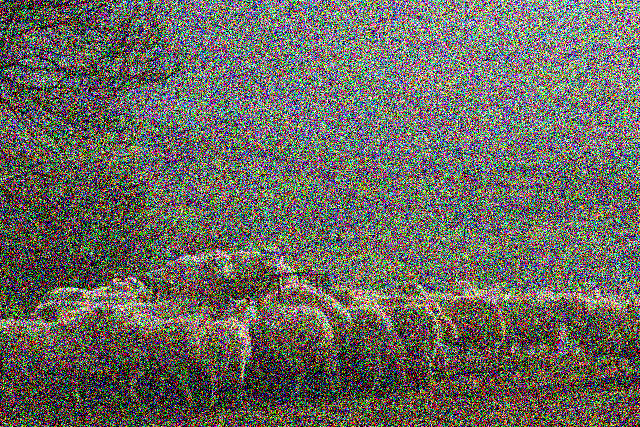} \\
          \small Original & \small Target & \small Context & \small Image \\[6pt]
          
          \multicolumn{4}{c}{%
              \includegraphics[height=0.17\textwidth]{Images/salience/low_salient_traffic.jpg} \hspace{4pt}         
              \includegraphics[height=0.17\textwidth]{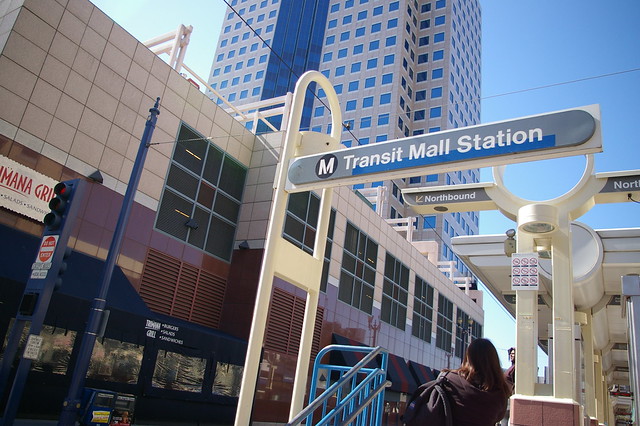} \hspace{4pt}         
              \includegraphics[height=0.17\textwidth]{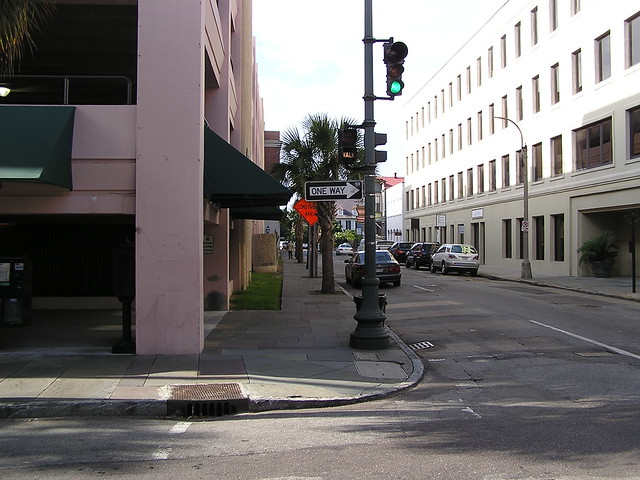}%
          } \\
          \multicolumn{4}{c}{%
              \small\hspace{4pt} Low salience \hspace{28pt} Medium salience \hspace{28pt} High salience%
          } \\
      \end{tabular}
      \caption{Top row: Gaussian noise applied to different regions of the image: the target category only (``sheep"), the
      surrounding context only, or the entire image. Bottom row: the concept ``traffic light" presented at three levels of visual salience. The images are credited to (from top right): jonbgem (license: CC-BY-NC), Henry de Saussure Copeland (license: CC-BY-NC), Frederick Dennstedt (license: CC-BY-SA), appelogen.be (license: CC-BY-NC).}.
      \label{fig:input_uncertainty}
  \end{figure}
  \clearpage
\newpage

\subsection{Context effects}
\label{app:mscoco_context}
\begin{figure}[H]
      \centering
      \includegraphics[width=\textwidth]{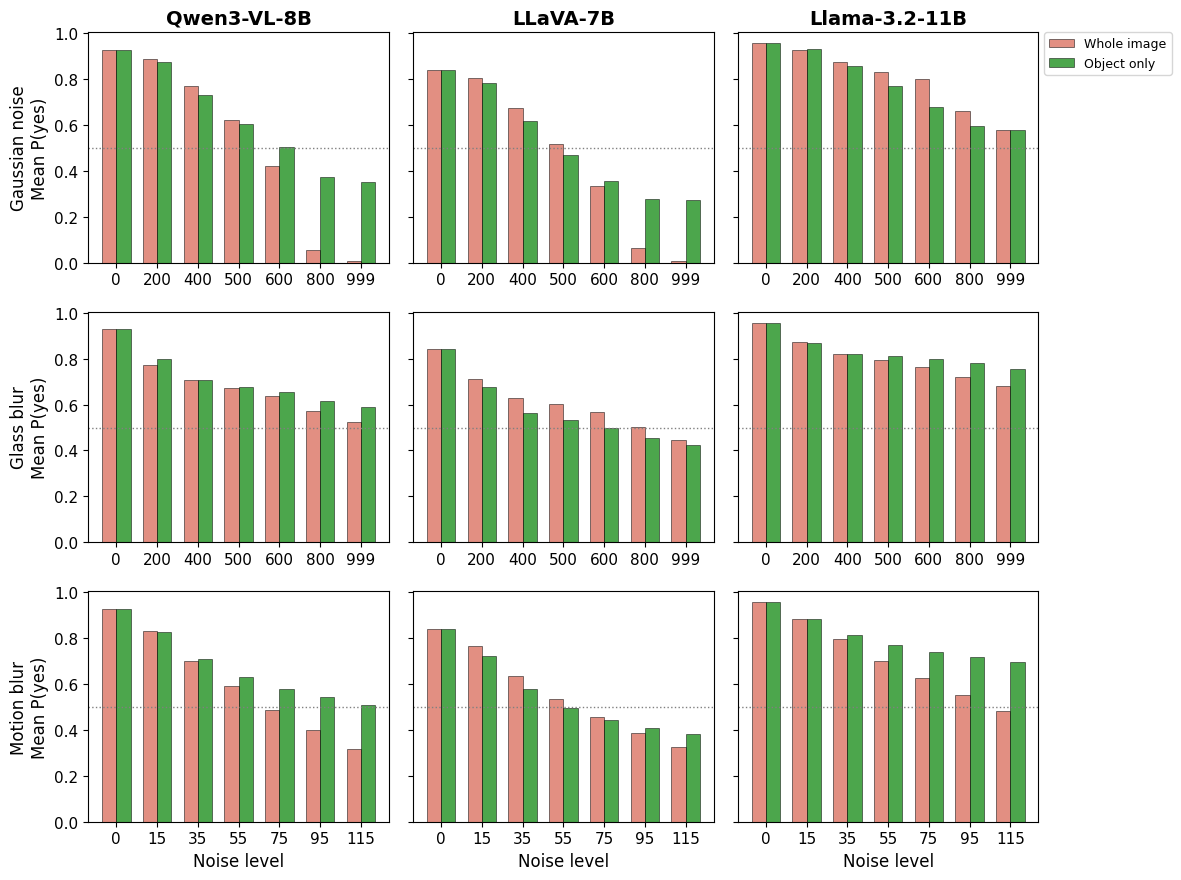}
      \caption{\textbf{Context information.} Final-layer mean $P(\text{yes})$ comparing whole-image perturbation (red) versus object-only perturbation (green) across noise levels, for the three models under consideration (columns) and our three noise types (rows), on MS-COCO images. When the context is not perturbed, $P(\text{yes})$ remains relatively high, even for high levels of noise. This is consistent with the non-random co-occurrence of objects and their surroundings: intact context provides prior expectations about which objects are likely present, partially compensating for the   
  degraded object region. The effect is most pronounced for LLaVA-7B and at high noise, where the gap between the two conditions widens substantially.}
      \label{fig:context_help}
  \end{figure}

\clearpage
\subsection{Impact of noise on MSCOCO responses}
\label{app:mscoco_noise_sweep}

To verify that image perturbations with noise meaningfully affect model outputs, we conducted the following experiment. We sampled 50  of 80 MS-COCO categories and selected 20 images per category from the train split, yielding approximately 1,000 positive image-category pairs. For each pair and noise level, we applied the perturbation to either the whole image or the object bounding box only, then queried LLaVA-7B with the prompt ``Is there a {category} in the image? Answer yes or no'' and recorded the first-token log-probabilities over the yes/no vocabulary. We report $P(\text{yes})$, computed by normalizing the logits for the ''yes'' and ''no'' tokens into a probability via softmax.

\begin{table}[H]                                                 
  \centering                                                             
  \label{tab:noise_comparison}                                            
  \begin{tabular}{ll*{7}{c}}                                              
  \toprule                                                                
  Perturbation & Region & \multicolumn{7}{c}{Noise Level} \\              
  \midrule                                                                
  \rowcolor{gray!20}                                        
  & & 0 & 200 & 400 & 500 & 600 & 800 & 999 \\
  \cmidrule(lr){3-9}
  \multirow{2}{*}{Gaussian noise} & Whole image & .837 & .797 & .642 &
  .472 & .300 & .085 & .031 \\
  & Object only & .837 & .770 & .574 & .456 & .387 & .347 & .343 \\
  \midrule
  \rowcolor{gray!20}
  & & 0 & 200 & 400 & 500 & 600 & 800 & 999 \\
  \cmidrule(lr){3-9}
  \multirow{2}{*}{Glass blur} & Whole image & .837 & .686 & .608 & .571 &
  .540 & .476 & .414 \\
  & Object only & .837 & .652 & .542 & .507 & .479 & .443 & .418 \\
  \midrule
  \rowcolor{gray!20}
  & & 0 & 15 & 35 & 55 & 75 & 95 & 115 \\
  \cmidrule(lr){3-9}
  \multirow{2}{*}{Motion blur} & Whole image & .837 & .761 & .623 & .516 &
   .433 & .378 & .331 \\
  & Object only & .837 & .713 & .558 & .486 & .449 & .428 & .412 \\
  \bottomrule
  \end{tabular}
  \caption{Mean $P(\text{yes})$ across perturbation types, noise levels, and blur  
  regions (LLaVA-7B, positive images only, $n=1000$).  Gaussian and glass blur levels are standardized to a 0–999 range to enable direct comparison, while motion blur kernel size is 
  bounded by image dimensions, preventing the same standardization.}   
  \end{table}

\section{Ablating the number of queries}
\label{app:ablation_n_queries}

\begin{figure}[ht]
    \centering
    \includegraphics[width=0.8\textwidth]{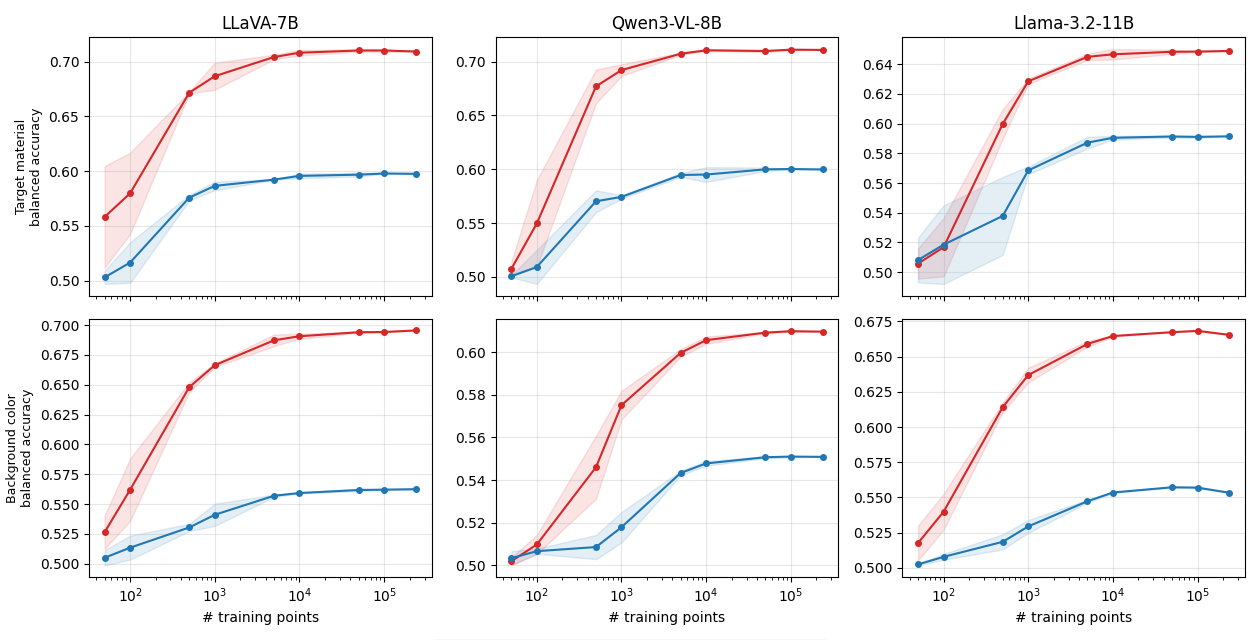}
    \caption{Performance of linear probes using the top-2L logits, on two attributes from the CLEVR dataset, as a function of training set size. Red lines use the true values of all included logits; blue lines only use the values of the top 20 logits in response to a query.}
    \label{fig:myimage}
\end{figure}

\section{LLM use disclosure}
Some of the code used in this work was generated by an LLM.

\applefootnote{ \textcolor{textgray}{\sffamily Apple and the Apple logo are trademarks of Apple Inc., registered in the U.S. and other countries and regions.}}

\end{document}